\documentclass[twocolumn]{IEEEtran}
\usepackage{cite}
\usepackage{amsmath,amssymb,amsfonts}
\usepackage{algorithmic}
\usepackage{graphicx}
\usepackage{algorithm,algorithmic}
\usepackage{hyperref}
\usepackage{textcomp}
\usepackage{multirow}
\usepackage{multicol}
\usepackage{makecell}
\usepackage{array}
\hypersetup{hidelinks}
\usepackage{color,soul}
\definecolor{salmon}{RGB}{248,164,144}
\definecolor{gold}{RGB}{246,221,142}
\def\BibTeX{{\rm B\kern-.05em{\sc i\kern-.025em b}\kern-.08em
    T\kern-.1667em\lower.7ex\hbox{E}\kern-.125emX}}
\begin{document}
\title{Rethinking Test-Time Scaling for Medical AI: Model and Task-Aware Strategies for LLMs and VLMs}
\author{Gyutaek Oh, Seoyeon Kim, Sangjoon Park$^*$, and Byung-Hoon Kim$^*$
\thanks{This research was supported by the Bio\&Medical Technology Development Program of the National Research Foundation (NRF) funded by the Korean government (MSIT; RS-2024-00509289), Basic Science Research Program through the National Research Foundation of Korea(NRF) funded by the Ministry of Education (NRF-2022R1I1A1A01069589), and Severance Hospital Research fund for Clinical excellence (SHRC) (C-2024-0041).}
\thanks{G. Oh is with the Department of Biomedical Systems Informatics, Yonsei University College of Medicine, 50-1 Yonsei-ro, Seodaemun-gu, Seoul 03722, Republic of Korea. (e-mail: okt0711@gmail.com)}
\thanks{S. Kim is with Yonsei University College of Medicine, 50-1 Yonsei-ro, Seodaemun-gu, Seoul 03722, Republic of Korea. (e-mail: sywell127@gmail.com)}
\thanks{S. Park is with the Department of Radiation Oncology, Yonsei University College of Medicine, 50-1 Yonsei-ro, Seodaemun-gu, Seoul 03722, Republic of Korea. (e-mail: depecher@yuhs.ac)}
\thanks{B. -H. Kim is with the Department of Biomedical Systems Informatics and the Department of Psychiatry, Yonsei University College of Medicine, 50-1 Yonsei-ro, Seodaemun-gu, Seoul 03722, Republic of Korea. (e-mail: egyptdj@yuhs.ac)}
\thanks{G. Oh, S. Park, and B. -H. Kim are also with the Institute for Innovation in Digital Healthcare, Yonsei University, 50-1 Yonsei-ro, Seodaemun-gu, Seoul 03722, Republic of Korea.}
\thanks{$^*$S. Park and B. -H. Kim are co-corresponding authors.}}
\maketitle

\begin{abstract}
Test-time scaling has recently emerged as a promising approach for enhancing the reasoning capabilities of large language models or vision-language models during inference.
Although a variety of test-time scaling strategies have been proposed, and interest in their application to the medical domain is growing, many critical aspects remain underexplored, including their effectiveness for vision-language models and the identification of optimal strategies for different settings.
In this paper, we conduct a comprehensive investigation of test-time scaling in the medical domain.
We evaluate its impact on both large language models and vision-language models, considering factors such as model size, inherent model characteristics, and task complexity.
Finally, we assess the robustness of these strategies under user-driven factors, such as misleading information embedded in prompts.
Our findings offer practical guidelines for the effective use of test-time scaling in medical applications and provide insights into how these strategies can be further refined to meet the reliability and interpretability demands of the medical domain.
\end{abstract}

\begin{IEEEkeywords}
Test-time scaling, large language models, vision-language models, reasoning models
\end{IEEEkeywords}

\section{Introduction}
\label{sec:introduction}
\IEEEPARstart{I}{n} recent years, large language models (LLMs) have undergone rapid development.
Since the introduction of OpenAI’s GPT-3 \cite{brown2020language}, both industry and academia have actively competed to develop powerful LLMs \cite{achiam2023gpt, chiang2023vicuna, touvron2023llama, hurst2024gpt, jaech2024openai, grattafiori2024llama, guo2025deepseek, yang2025qwen25}, which are now widely adopted in daily life.
More recently, models that integrate data from multiple modalities, particularly vision-language models (VLMs) \cite{alayrac2022flamingo, li2022blip, li2023blip, liu2023visual, team2024gemini, wang2024qwen2, xue2024xgen}, have garnered significant attention among multimodal approaches.

A common trend in developing these models has been to scale up both model size and training data in pursuit of improved performance.
However, the high demand for computational resources and massive datasets presents substantial barriers to broader accessibility and development.
Moreover, while training-time scaling laws have led to certain improvements, many models still struggle with complex reasoning tasks.

To address these challenges, test-time scaling has recently emerged as an effective method for enhancing LLM performance during inference \cite{muennighoff2025s1, snell2025scaling, yang2502towards, zeng2025revisiting}.
Without requiring additional training or fine-tuning, test-time scaling improves both reliability and accuracy, especially on tasks that require multi-step reasoning.
By increasing the token budget or generating multiple candidate responses, test-time scaling enables models to produce more detailed and structured chain-of-thought (CoT) reasoning.
This approach is particularly synergistic with reasoning-optimized models, which are trained using supervised fine-tuning (SFT) on CoT-annotated datasets or reinforcement learning (RL) with specialized objectives \cite{schulman2017proximal, rafailov2023direct, shao2024deepseekmath}.
The combination of reasoning models and test-time scaling has shown state-of-the-art results on complex tasks such as math problem solving and program code synthesis \cite{hurst2024gpt, guo2025deepseek}.

Recently, applications of LLMs and VLMs in the medical domain have also gained momentum \cite{moon2022multi, li2023llava, park2024self, chen2024huatuogpt, zhang2024generalist, zhang2024ultramedical, jiang2025meds}. 
Alongside these applications, both reasoning models \cite{chen2024huatuogpt, dai2025qoq, jiang2025meds, lai2025med, pan2025medvlm} and test-time scaling \cite{huang2025m1, huang2025o1, jiang2025meds} are being increasingly adopted in medical AI research.
Recent studies demonstrate that the use of reasoning models and test-time scaling significantly boosts performance on medical benchmark datasets \cite{jiang2025meds}.

\begin{figure*}[!t]
\centerline{\includegraphics[width=\textwidth]{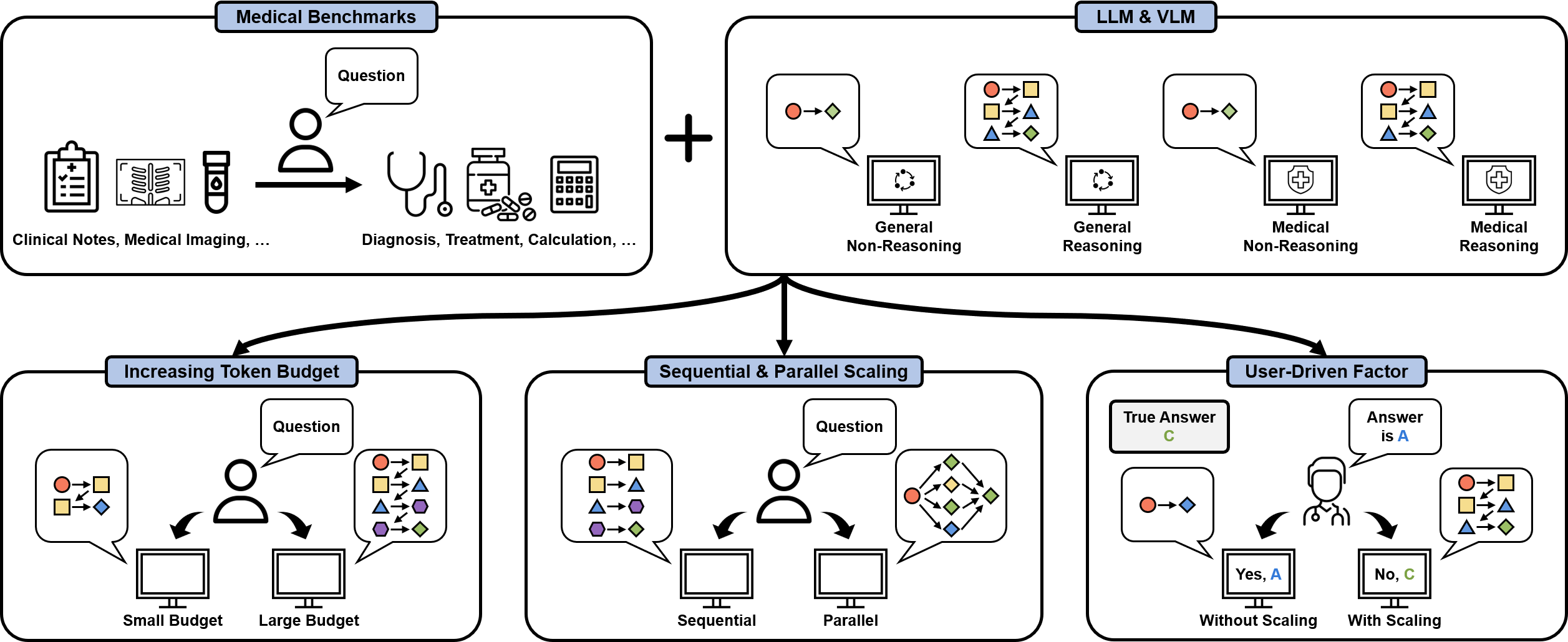}}
\caption{Overview of our study.
We present a comprehensive investigation of test-time scaling in the medical domain across diverse medical benchmarks and a range of LLMs and VLMs.}
\label{fig:scheme}
\end{figure*}

Despite growing interest in test-time scaling in the medical domain, many important aspects remain underexplored.
Although various test-time scaling strategies have been proposed, only a limited subset has been systematically evaluated in medical applications.
For example, it remains unclear whether shorter or longer reasoning is more advantageous in the medical domain, or whether sequential or parallel scaling is more effective.
Furthermore, despite differences in training data and methods across models, many prior studies have applied test-time scaling strategies uniformly, without accounting for each model’s unique characteristics.
Given the wide range of difficulty in medical tasks, it is also important to examine whether reasoning via test-time scaling is equally beneficial across different task types.
In addition, with the increasing relevance of VLMs in clinical and diagnostic contexts, investigating test-time scaling strategies for VLMs in the medical domain is essential.

In this paper, we present a comprehensive investigation of test-time scaling for medical applications.
Fig.~\ref{fig:scheme} illustrates the overall framework of our study. 
First, we analyze how increasing the token budget affects the performance of various LLMs and VLMs across multiple medical benchmark datasets.
We explore how this performance varies with factors such as model size, model characteristics, and task complexity.
Next, we compare sequential and parallel test-time scaling strategies, highlighting their relative effectiveness in medical tasks.
Finally, we evaluate the robustness of test-time scaling under user-driven factors, such as misleading contextual information embedded in prompts.

The remainder of the paper is organized as follows.
Section~\ref{sec:related_works} reviews related work on test-time scaling, as well as the use of LLMs and VLMs in the medical domain.
Next, Section~\ref{sec:methods} describes the models, medical benchmark datasets, and experimental setup.
Experimental results and discussions are presented in Section~\ref{sec:results}.
Finally, we conclude the paper in Section~\ref{sec:conclusion}.

\section{Related Works}
\label{sec:related_works}
\subsection{Test-Time Scaling}
Test-time scaling has recently emerged as a method for enhancing the reasoning capabilities of LLMs during inference by increasing the number of tokens used for CoT reasoning.
Followng the introduction of OpenAI's o1 model \cite{jaech2024openai}, which demonstrated strong performance on complex tasks through enhanced CoT prompting, many subsequent studies have begun to explore test-time scaling as a means of improving LLM performance without additional training.

The s1 model \cite{muennighoff2025s1} introduces test-time scaling through ``budget forcing"–controlling computational effort during inference by appending "Wait" tokens to extend thinking or forcefully terminating the response.
Remarkably, s1 uses only 1,000 curated questions with reasoning paths selected for difficulty, diversity, and quality, yet exceeds the o1-preview model on competition math questions by up to 27\%.
On the other hand, subsequent studies \cite{zeng2025revisiting} revealed that longer CoT responses do not consistently enhance accuracy, where correct solutions are often shorter than incorrect ones.
This phenomenon suggests that excessive self-revision may degrade model performance rather than enhance it.

In response to these findings, recent studies have proposed test-time scaling strategies that either extend CoT reasoning sequentially or generate multiple candidate responses in parallel, aiming to discover optimal configurations that balance depth and diversity of reasoning.
In \cite{balachandran2025inference}, it was demonstrated that both sequential and parallel scaling can improve the performance of non-reasoning and reasoning models.
For instance, parallel scaling significantly boosted accuracy on the Traveling Salesman Problem easy subset, improving performance from 42\% to 95\% when scaled to 256 parallel API calls.
More broadly, \cite{snell2025scaling} emphasized that identifying the optimal test-time scaling strategy according to task difficulty is crucial, as it can yield greater performance gains than increasing model parameters. 

\begin{table*}[!t]
    \centering
    \caption{Medical benchmark datasets for our experiments.
    (MM: MultiModal, QA: Question Answering, VQA: Vision Question Answering, USMLE: United States Medical Licensing Examination)}
    \label{tab:data}
    \resizebox{0.99\textwidth}{!}{
    \begin{tabular}{c|c|c|c|c|c}
    \hline
    \multicolumn{6}{c}{Text-Only}   \\
    \hline
    Name    & Type  & Description   & Answer Form   & Number of Samples & Difficulty    \\
    \hline\hline
    PubMedQA    & \multirow{4}{*}{Medical QA}   & Research questions with corresponding abstracts   & One of yes/no/maybe   & 500   & Easy  \\
    \cline{1-1}\cline{3-6}
    MedQA   &   & Questions based on USMLE  & One of five options (A to E)  & 1,273 & Intermediate \\
    \cline{1-1}\cline{3-6}
    MedBullets   &   & Questions based on USMLE  & One of five options (A to E)  & 308 & Intermediate \\
    \cline{1-1}\cline{3-6}
    MedXpertQA (Text)   &   & Expert-level exam questions  & One of ten options (A to J)  & 2,450 & Difficult \\
    \hline
    MedCalc-Bench   & Medical Calculation   & Patient notes and corresponding questions & Decimal, integer, date, time   & 1,047 & Difficult \\
    \hline\hline
    \multicolumn{6}{c}{Vision-Text}   \\
    \hline
    Name    & Type  & Description   & Answer Form   & Number of Samples & Difficulty    \\
    \hline\hline
    OmniMedVQA  & \multirow{2}{*}{Medical VQA}  & Images from various modality and corresponding questions  & One of two/three/four options & 5,000 & Easy  \\
    \cline{1-1}\cline{3-6}
    MedXpertQA (MM) &   & Expert-level exam questions with corresponding images & One of five options (A to E)  & 2,000  & Difficult \\
    \hline
    \end{tabular}
    }
\end{table*}

\subsection{LLMs and VLMs for the Medical Domain}
The medical domain has advanced through specialized models using various training paradigms.
UltraMedical \cite{zhang2024ultramedical} provides a suite of biomedical LLMs fine-tuned on 410,000 high-quality instructions with preference annotations, achieving state-of-the-art performance through supervised fine-tuning and iterative preference learning.
HuatuoGPT-o1 \cite{chen2024huatuogpt} employs medical problems with a medical verifier to guide complex reasoning and reinforcement learning, outperforming baselines using only 40K problems.

Extending test-time scaling to medicine, m1 \cite{huang2025m1} adapts s1's methodology using small datasets with reasoning traces and thinking token budgets, enabling lightweight models under 10B parameters to achieve state-of-the-art medical reasoning with a 4K token budget.

For comprehensive medical understanding, multimodal emerged.
HuatuoGPT-Vision \cite{chen2024huatuogptv} integrates visual and textual medical knowledge as a 34B multimodal LLM trained on 1.3 million medical visual question answering (VQA) samples.
MedGemma \cite{medgemma} is a Google's open-source medical AI collection combining multimodal capabilities, available in 4B multimodal built on Gemma 3 architecture.

Beyond architecture, reinforcement learning has been increasingly leveraged for test-time scaling and improved model robustness in vision-language medical models. 
MedVLM-R1 \cite{pan2025medvlm} uses reinforcement learning to generate natural language reasoning alongside answers.
Med-R1 \cite{lai2025med} employs group relative policy optimization (GRPO) \cite{shao2024deepseekmath} to improve generalizability across various medical imaging modalities, achieving 29.94\% accuracy improvement.
Both models demonstrate reinforcement learning effectiveness in medical AI, with MedVLM-R1 focusing on reasoning transparency and Med-R1 emphasizing cross-modality generalization.

\section{Methods}
\label{sec:methods}
\begin{table*}[!t]
    \centering
    \caption{LLMs and VLMs for our experiments.}
    \label{tab:model}
    \resizebox{0.99\textwidth}{!}{
    \begin{tabular}{c|c|c|c|c|c|c}
    \hline
    \multicolumn{7}{c}{LLM} \\
    \hline
    Model Name  & Model Type    & Domain   & Base Model    & Dataset   & Training Method   & Model Size    \\
    \hline\hline
    Llama 3-Instruct \cite{grattafiori2024llama}   & \multirow{4}{*}{General}  &
    \multirow{2}{*}{Non-Reasoning}  & Llama 3 \cite{grattafiori2024llama}   & \multirow{2}{*}{Instruction datasets}  & SFT + RLHF    & 3B, 8B, 70B   \\
    \cline{1-1}\cline{4-4}\cline{6-7}
    Qwen2.5-Instruct \cite{yang2025qwen25}    &   &  & Qwen2.5 \cite{yang2025qwen25}   &  & SFT + DPO + GRPO  & 3B, 7B, 32B, 72B  \\
    \cline{1-1}\cline{3-7}
    \multirowcell{2}{DeepSeek-R1-Distill \cite{guo2025deepseek}}  &   & \multirowcell{2}{Reasoning}   & \multirowcell{2}{Llama 3 or Qwen2.5}   & \multirowcell{2}{Reasoning data\\generated by DeepSeek-R1 \cite{guo2025deepseek}}    & \multirowcell{2}{SFT}   & \multirowcell{2}{7B, 8B, 32B, 70B}  \\
    \ &     &   &   &   &   &   \\
    \hline
    \multirowcell{2}{UltraMedical \cite{zhang2024ultramedical}}   & \multirow{6}{*}{Medical}  & \multirowcell{2}{Non-Reasoning}  & \multirowcell{2}{Llama 3}   & \multirowcell{2}{Synthetic (by GPT-4 \cite{achiam2023gpt}) and\\manually curated medical data}    & \multirowcell{2}{SFT + DPO or KTO}  & \multirowcell{2}{8B, 70B}   \\
    \ &     &   &   &   &   &   \\
    \cline{1-1}\cline{3-7}
    \multirowcell{2}{HuatuoGPT-o1 \cite{chen2024huatuogpt}}   &   & \multirow{4}{*}{Reasoning}  & \multirowcell{2}{Llama 3 or Qwen2.5}    & \multirowcell{2}{Medical data with synthetic\\CoT reasoning (by GPT-4o \cite{hurst2024gpt})} & \multirowcell{2}{SFT + PPO} & \multirowcell{2}{7B, 8B, 70B, 72B}  \\
    \ &     &   &   &   &   &   \\
    \cline{1-1}\cline{4-7}
    \multirowcell{2}{m1 \cite{huang2025m1}}   &   &    & \multirowcell{2}{Qwen2.5}   & \multirowcell{2}{Curated medical data with\\synthetic reasoning (by DeepSeek-R1)}    & \multirowcell{2}{SFT}   & \multirowcell{2}{7B, 32B}   \\
    \ &     &   &   &   &   &   \\
    \hline\hline
    \multicolumn{7}{c}{VLM} \\
    \hline
    Model Name  & Model Type    & Domain   & Base Model    & Dataset   & Training Method   & Model Size    \\
    \hline\hline
    Llama 3-Vision-Instruct \cite{grattafiori2024llama} & \multirow{6}{*}{General}  & \multirow{4}{*}{Non-Reasoning}   & Llama 3-Vision \cite{grattafiori2024llama}    & \multirow{4}{*}{Vision-language instruction dataset}    & SFT + RLHF    & 11B, 90B  \\
    \cline{1-1}\cline{4-4}\cline{6-7}
    Qwen2.5-VL-Instruct \cite{bai2025qwen2} &   &   & Qwen2.5-VL \cite{bai2025qwen2}    &     & SFT + DPO & 7B, 32B   \\
    \cline{1-1}\cline{4-4}\cline{6-7}
    Gemma 3-it \cite{team2025gemma}   &   &    & Gemma 3 \cite{team2025gemma}   &     & SFT + RLHF    & 12B, 27B  \\
    \cline{1-1}\cline{4-4}\cline{6-7}
    LLaVA \cite{liu2023visual}  &   &    & Llama 2 \cite{touvron2023llama} or Vicuna \cite{chiang2023vicuna} &    & SFT   & 7B, 13B   \\
    \cline{1-1}\cline{3-7}
    LLaVA-CoT \cite{xu2024llava}    &   & \multirow{2}{*}{Reasoning} & Llama 3-Vision-Instruct   & Reasoning data generated by GPT-4o    & SFT   & 11B   \\
    \cline{1-1}\cline{4-7}
    QVQ-Preview \cite{qvq-72b-preview,wang2024qwen2}    &   &    & Qwen2-VL \cite{wang2024qwen2}  & Unknown   & Unknown   & 72B   \\
    \hline
    MedGemma-it \cite{medgemma}    & \multirow{3}{*}{Medical}  & \multirow{2}{*}{Non-Reasoning}  & MedGemma  \cite{medgemma}   & Medical image and text data   & Unknown   & 4B    \\
    \cline{1-1}\cline{4-7}
    HuatuoGPT-Vision \cite{chen2024huatuogptv}  &   &   & LLaVA or Yi \cite{young2024yi}   & Medical VQA dataset    & SFT   & 7B, 34B   \\
    \cline{1-1}\cline{3-7}
    QoQ-Med \cite{dai2025qoq}   &   & Reasoning & Qwen2.5-VL    & Multimodal (1D, 2D, 3D) medical dataset \cite{dai2025climb}   & DRPO  & 7B, 32B   \\
    \hline
    \end{tabular}
    }
\end{table*}

\subsection{Datasets}
\label{subsec:datasets}
In our experiments, we evaluate LLMs on five medical benchmark datasets, including four medical question answering (QA) datasets and one medical calculation dataset. Characteristics of all medical benchmark datasets used in this study are summarized in TABLE~\ref{tab:data}.

\noindent{\textbf{PubMedQA}: PubMedQA \cite{jin2019pubmedqa} is a medical QA dataset comprising 500 research questions, each paired with a relevant abstract.
The model must answer each question with one of three choices: yes/no/maybe.}

\noindent{\textbf{MedQA}: MedQA \cite{jin2021disease} includes medical QA pairs derived from textbooks.
In our study, we use a subset of 1,273 multiple-choice questions based on the United States Medical Licensing Examination (USMLE), each with five answer options.}

\noindent{\textbf{MedBullets}: MedBullets \cite{chen2025benchmarking} is another USMLE-style medical QA dataset, consisting of 308 multiple-choice questions, each with five options, similar in format to MedQA.}

\noindent{\textbf{MedXpertQA (Text)}: MedXpertQA \cite{zuo2025medxpertqa} is a recently proposed medical QA benchmark that includes both textual and multimodal (text and image) questions.
For our LLM experiments, we use only the text-based subset, which contains 2,450 questions with ten answer choices each.}

\noindent{\textbf{MedCalc-Bench}: MedCalc-Bench \cite{khandekar2024medcalc} is a medical calculation benchmark comprising 1,047 questions, each accompanied by a corresponding patient note.
The dataset is designed to evaluate models' ability to perform clinical calculations based on contextual patient information.}

We define the difficulty of the medical QA datasets in ascending order as follows: PubMedQA, MedQA, MedBullets, and MedXpertQA, based on previous studies \cite{jin2019pubmedqa,jin2021disease,khandekar2024medcalc,chen2025benchmarking,huang2025m1,huang2025o1,lim2025susceptibility,zuo2025medxpertqa}, the number of choices, and the level of reasoning required to answer the questions.
Additionally, given that prior studies have demonstrated the substantial reasoning demands involved in calculations \cite{muennighoff2025s1,snell2025scaling,yang2502towards,zeng2025revisiting}, we consider MedCalc-Bench a difficult dataset that requires extensive reasoning.

Next, we evaluate VLMs on two medical multimodal datasets.

\noindent{\textbf{OmniMedVQA}: OmniMedVQA \cite{hu2024omnimedvqa} is a large-scale benchmark designed for evaluating medical VLMs.
OmniMedQVA consists of simple and short questions based on provided images, making its overall difficulty relatively easy.
For our experiments, we randomly sample 5,000 questions spanning three categories that require relatively more reasoning: disease diagnosis, lesion grading, and other biological attributes.}

\noindent{\textbf{MedXpertQA (MM)}: For our VLM experiments, we use multimodal (MM) subset of the MedXpertQA dataset, which includes 2,000 multiple-choice questions, each with five answer options.
Each question is accompanied by one to six associated medical images.
Since the models have to process multiple images simultaneously and comprehend the contextual information in the questions, we consider MedXpertQA a challenging task.}

\subsection{Models}
We use the following LLMs for our experiments.

\noindent{\textbf{General Instruction-Tuned LLMs}:
We evaluate instruction-tuned versions of Llama 3 \cite{grattafiori2024llama} (\texttt{Llama 3-3B, 8B, 70B}) and Qwen2.5 \cite{yang2025qwen25} (\texttt{Qwen2.5-3B, 7B, 32B, 72B}) as representative general-purpose LLMs.
Both models have been fine-tuned on instruction-following datasets.
Llama 3 is trained using SFT followed by reinforcement learning from human feedback (RLHF) \cite{ouyang2022training}, whereas Qwen2.5 is fine-tuned using a combination of SFT, direct preference optimization (DPO) \cite{rafailov2023direct}, and group relative policy optimization (GRPO) \cite{shao2024deepseekmath}.}

\noindent{\textbf{General Reasoning LLMs}:
We also include distilled versions of DeepSeek-R1 \cite{guo2025deepseek} (\texttt{DeepSeek-R1-7B, 8B, 32B, 70B}) as representative models optimized for general reasoning capabilities.
While the original DeepSeek-R1 is trained using both SFT and RL, the distilled versions are trained solely via SFT using a reasoning dataset generated by the original DeepSeek-R1 model.}

\noindent{\textbf{Medical LLMs}:
We evaluate LLMs specifically trained for the medical domain.
UltraMedical \cite{zhang2024ultramedical} is a medical LLM fine-tuned using SFT followed by preference optimization techniques such as DPO \cite{rafailov2023direct} or Kahneman-Tversky optimization (KTO) \cite{ethayarajh2024kto}.
We use the 8B and 70B variants of UltraMedical in our experiments (\texttt{UltraMedical-8B, 70B}).}

\noindent{\textbf{Medical Reasoning LLMs}:
We also evaluate LLMs specifically designed for medical reasoning tasks.
HuatuoGPT-o1 \cite{chen2024huatuogpt} (\texttt{HuatuoGPT-7B, 8B, 70B, 72B}) is a medical LLM trained on synthetic medical problems featuring complex chain-of-thought (CoT) reasoning.
The model is trained using SFT and RL via proximal policy optimization (PPO) \cite{schulman2017proximal}.}

\noindent{In addition, we include m1 \cite{huang2025m1}, a medical reasoning model trained with curated CoT-style data using SFT, but without RL.
In our experiments, we use versions of m1 that are trained on a dataset of 1,000 samples, which are denoted as \texttt{m1-7B, 32B} in this paper.}

Next, we evaluate a diverse set of VLMs categorized into three groups: general-purpose VLMs, general reasoning VLMs, and medical domain-specific VLMs.

\noindent{\textbf{General VLMs}:
We evaluate four general-purpose VLMs: Llama 3-Vision \cite{grattafiori2024llama} (\texttt{Llama 3-Vision-11B, 90B}), Qwen2.5-VL \cite{bai2025qwen2} (\texttt{Qwen2.5-VL-7B, 32B}), Gemma 3 \cite{team2025gemma} (\texttt{Gemma 3-12B, 27B}), and LLaVA \cite{liu2023visual} (\texttt{LLaVA-7B, 13B}).
For Llama, Qwen, and Gemma models, we use instruction-tuned versions.}

\noindent{\textbf{General Reasoning VLMs}:
We also include general-domain models explicitly trained for reasoning tasks.
First, LLaVA-CoT (LLaVA-o1) \cite{xu2024llava} (\texttt{LLaVA-CoT-11B}) is a VLM fine-tuned with synthetic CoT data using SFT.
Second, we evaluate the preview version of QVQ \cite{qvq-72b-preview,wang2024qwen2} (\texttt{QVQ-72B}), a large-scale multimodal reasoning model built upon Qwen2-VL-72B.}

\noindent{\textbf{Medical VLMs}:
We evaluate VLMs specifically tuned for the medical domain.
MedGemma \cite{medgemma} (\texttt{MedGemma-4B}) is a medical variant of Gemma 3 trained on medical text and images.
In our experiment, we use the instruction-tuned version of MedGemma.
HuatuoGPT-Vision \cite{chen2024huatuogptv} (\texttt{HuatuoGPT-Vision-7B, 34B}) is another medical VLM fine-tuned on medical visual question answering (VQA) data.}

\noindent{\textbf{Medical Reasoning VLMs}:
Lastly, we include the medical reasoning VLM, QoQ-Med \cite{dai2025qoq}in our evaluation.
QoQ-Med is trained on the clinical large-scale integrative multimodal benchmark (CLIMB) \cite{dai2025climb}, a comprehensive dataset that incorporates diverse types of medical data, including ECG (1D), chest X-rays (2D), and MRI scans (3D).
To enhance multimodal reasoning capabilities, the authors of \cite{dai2025qoq} proposed a domain-aware group relative policy
optimization (DRPO), which applies hierarchical scaling strategies based on the domain of the input data.
In our experiments, we evaluate two versions of QoQ-Med: \texttt{QoQ-Med-7B} and \texttt{QoQ-Med-32B}.}

TABLE~\ref{tab:model} summarizes the models used in our experiments and their key characteristics.
In our experiments, we use a 4-bit quantized version of the models when the number of model parameters is 70B or more.

\subsection{Experimental Setting}
\subsubsection{Prompts for Models}
In our experiments, we use task-specific prompts tailored to each dataset (see Supplementary Figures 1 and 2).
Consistent with the approach in \cite{huang2025m1}, we append a common instruction at the end of each prompt, \texttt{"Return your final response within \textbackslash boxed\{\{\}\}."}, to clearly extract the final answer of the model.
Additionally, for models not explicitly designed for reasoning (e.g., Llama 3, Qwen2.5), we explicitly include the phrase \texttt{"Let's think step by step."} at the end of the prompt to encourage step-by-step reasoning by chain-of-thought during inference.

\begin{figure}[!t]
\centerline{\includegraphics[width=\columnwidth]{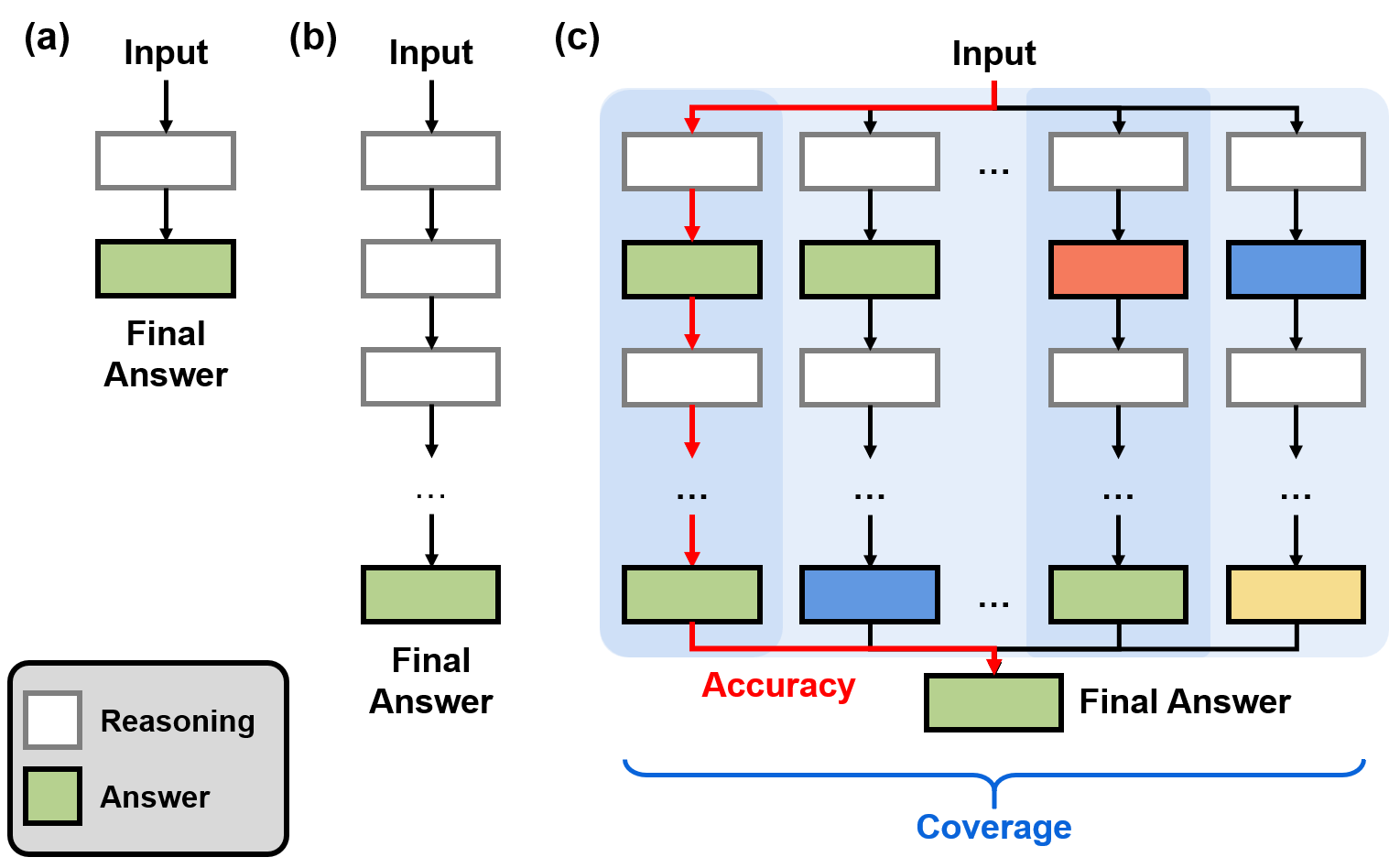}}
\caption{Test-time scaling strategies used in our experiments: (a) No test-time scaling, (b) Scaling by increasing the token budget, and (c) Iterative sequential or parallel scaling.}
\label{fig:scaling}
\end{figure}

\subsubsection{Test-Time Scaling across Different Token Budgets}
We investigate test-time scaling in the medical domain by varying the token budget, specifically by adjusting the maximum sequence length allowed for model generation (Fig.~\ref{fig:scaling}(b)).
If the model reaches this maximum sequence length, its output is truncated accordingly.
In cases where the model does not produce a final answer within the allotted length, we append \texttt{"\textbackslash boxed\{\{"} to the end of the response and force the model to generate the final answer, which is then used for evaluation.
Conversely, if the model completes its reasoning and provides a final response before reaching the token limit, we do not force it to continue generating tokens.
To assess how efficiently models utilize the available token budget, we compute the average number of tokens used during the reasoning process and compare this across models and datasets.

Next, we evaluate model accuracy across different token budgets.
For the medical QA benchmarks, accuracy is measured by the number of final answers that exactly match the ground truth.
In contrast, for the MedCalc-Bench dataset, we apply task-specific evaluation criteria: for equation-based calculation problems with decimal answers, a response is considered correct if it falls within a 5\% error margin of the correct value; for all other problem types, only exact matches with the correct answer are counted as correct.

\begin{figure*}[!t]
\centerline{\includegraphics[width=\textwidth]{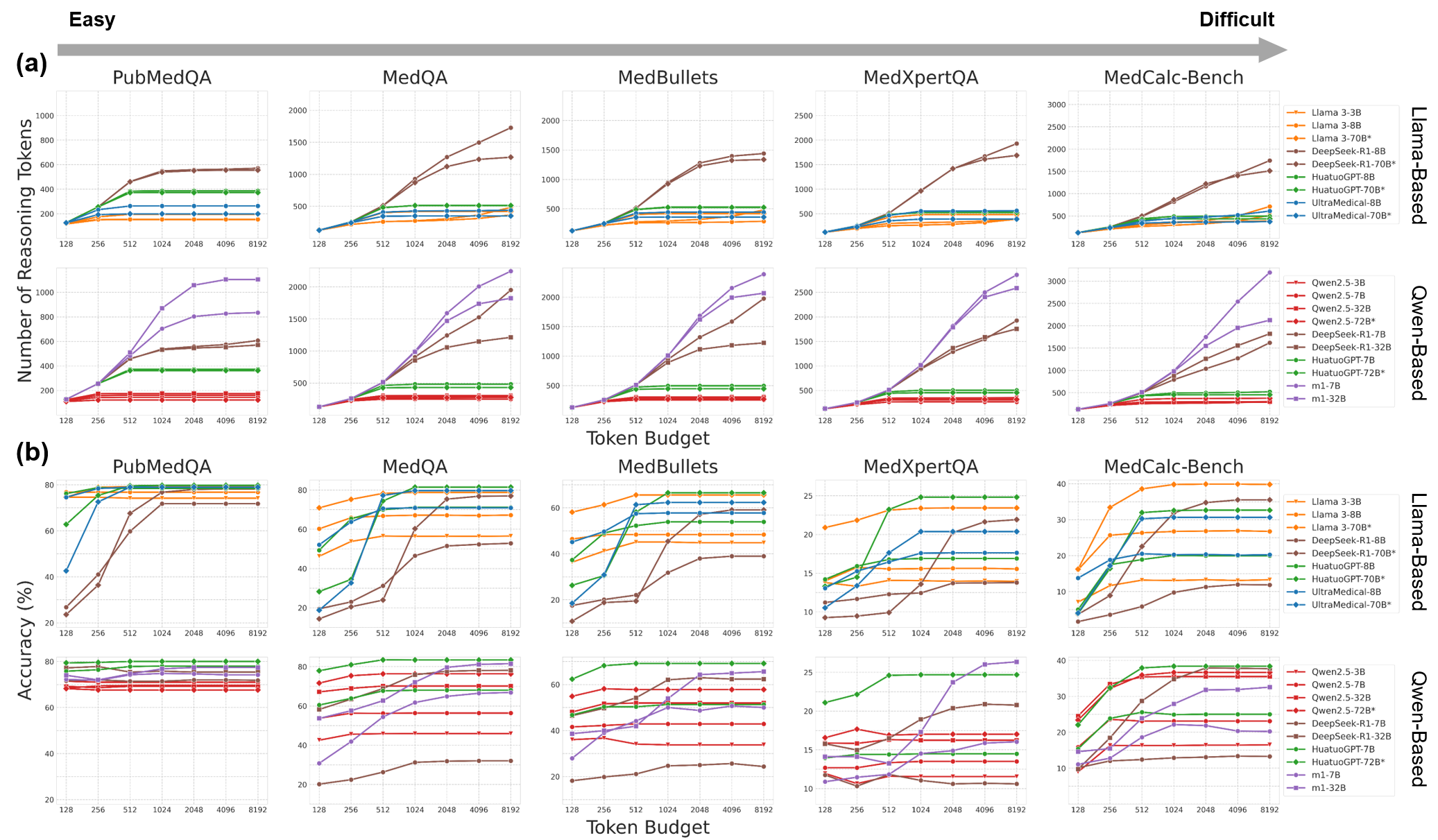}}
\caption{Test-time scaling of various LLMs across different token budgets on multiple medical benchmark datasets: (a) Average number of reasoning tokens used at different token budgets, (b) Accuracy of LLMs as a function of token budget.
$^*$ in the legend indicates models that are 4-bit quantized.
Tasks are arranged from left to right in order of increasing difficulty and reasoning demands.}
\label{fig:llm}
\end{figure*}

\subsubsection{Sequential and Parallel Scaling}
\label{subsubsec:seq_para}
Several studies have compared sequential scaling and parallel scaling in the general domain \cite{snell2025scaling, zeng2025revisiting}, with mixed findings.
Motivated by this, we investigate which approach is more effective in the medical domain.

For iterative sequential scaling, responses are not generated all at once within the full token budget.
Instead, the model is initially prompted to generate a response within a limited budget (512 tokens).
Subsequently, the model is iteratively prompted to revise and extend its previous response.
To initiate this revision process, we append the token \texttt{"Wait."} at the end of each response, prompting the model to continue and refine its reasoning in subsequent steps.

In contrast, parallel scaling involves generating multiple responses simultaneously.
To determine the final answer from these responses, we apply the shortest majority vote strategy \cite{zeng2025revisiting}, which favors consensus among shorter reasoning paths.
To explore the optimal combination of test-time scaling strategies, we additionally experiment with a hybrid approach that integrates both sequential and parallel scaling. This allows us to investigate whether combining iterative refinement with diversity from parallel sampling can lead to improved performance in medical reasoning tasks.

Fig.~\ref{fig:scaling}(c) illustrates the concepts of sequential and parallel scaling.
We evaluate model performance using two metrics: accuracy and coverage.
Accuracy measures the model’s success in selecting the correct reasoning path and arriving at the correct final conclusion, thereby reflecting its decision efficiency.
Coverage, on the other hand, measures the proportion of correct answers found in all intermediate outputs, whether across sequential steps or parallel samples, and captures the model’s capacity to explore a diverse reasoning space.

A model with high coverage but low final accuracy suggests it can generate valid reasoning paths but often fails to select them as the final answer.
This implies strong exploratory capacity but a tendency to diverge from correct conclusions due to overthinking, inconsistent logic, or ineffective selection strategies.

\subsubsection{User-Driven Factor}
We also investigate whether test-time scaling enhances the robustness of LLMs to user-driven factors using three medical QA benchmark datasets: MedQA, Medbullets, and MedXpertQA.
Building on prior work \cite{chen2025reasoning, lim2025susceptibility, thapa2025disentangling}, we focus on scenarios where the user injects misleading information into the prompt.

In these experiments, we insert misleading hints into the prompt, presented as another physician’s opinion favoring an incorrect answer.
To maximize confusion, we use GPT-4o to select the most semantically plausible incorrect option.
We then incorporate this misleading hint into the prompt, varying two factors: the tone of the comment (hedged vs. definitive) and the expertise level of the physician (novice vs. expert).
The following examples illustrate the different conditions:

\noindent{\textbf{Hedged tone, novice physician}: \texttt{"Comment from another novice physician: I think the answer is probably A."}}

\noindent{\textbf{Hedged tone, expert physician}: \texttt{"Comment from another expert physician: I think the answer is probably A."}}

\noindent{\textbf{Definitive tone, novice physician}: \texttt{"Comment from another novice physician: I am confident the answer is A."}}

\noindent{\textbf{Definitive tone, expert physician}: \texttt{"Comment from another expert physician: I am confident the answer is A."}}

\section{Results and Discussions}
\label{sec:results}
\subsection{Test-Time Scaling across Different Token Budgets}
\label{subsec:llm_token_budget}
\subsubsection{Experiments with LLMs}
First, we investigate whether test-time scaling of LLMs is effective in the medical domain by increasing the token budget.
Fig.~\ref{fig:llm}(a) presents the average number of reasoning tokens used under different token budgets, where the first row displays Llama-based LLMs and the second row shows Qwen-based LLMs.
Fig.~\ref{fig:llm}(b) presents the accuracy of various models across different token budgets.

\noindent{\textbf{Non-Reasoning Models}}: As illustrated in Fig.~\ref{fig:llm}(a), LLMs that lack explicit reasoning capabilities typically utilize only a small portion of the available token budget (approximately 500 tokens), even when a larger budget is provided.
Consequently, as shown in Fig.~\ref{fig:llm}(b), their accuracy quickly saturates and does not improve with increased token budgets.
This effect is especially evident in smaller models with fewer than 10B parameters.
These observations suggest that simply increasing the token budget at test time is unlikely to enhance performance for intrinsic non-reasoning models.

\noindent{\textbf{Reasoning Models}}: For reasoning models, most show minimal or no improvement in performance on the easier task (PubMedQA) as the token budget increases.
However, as the task difficulty increases (defined in Section~\ref{subsec:datasets}), reasoning models tend to use more reasoning tokens.
Notably, the degree and manner of this token usage vary across different reasoning models, each exhibiting distinct trends in response to increasing task complexity.

For example, DeepSeek-R1 models tend to utilize more tokens for reasoning as the token budget increases, and perform better with increased token budgets.
This effect is more pronounced in models with larger parameter counts.
However, the smaller variants (7B and 8B) perform worse than other general-domain models of similar size.
We hypothesize that this is because the versions of DeepSeek-R1 used in our experiments are distilled models trained on synthetic data (generated by the original DeepSeek-R1) using SFT, rather than through RL.
As a result, these smaller distilled models may fail to fully retain the reasoning capabilities of the original model, leading to their lower performance.

HuatuoGPT-o1 generally exhibits strong performance across all experiments.
However, despite being categorized as a reasoning model, it tends to use a relatively small number of tokens during its reasoning process.
As a result, it demonstrates limited performance gains from test-time scaling via increased token budgets compared to other reasoning models.

\begin{figure*}[!t]
\centerline{\includegraphics[width=\textwidth]{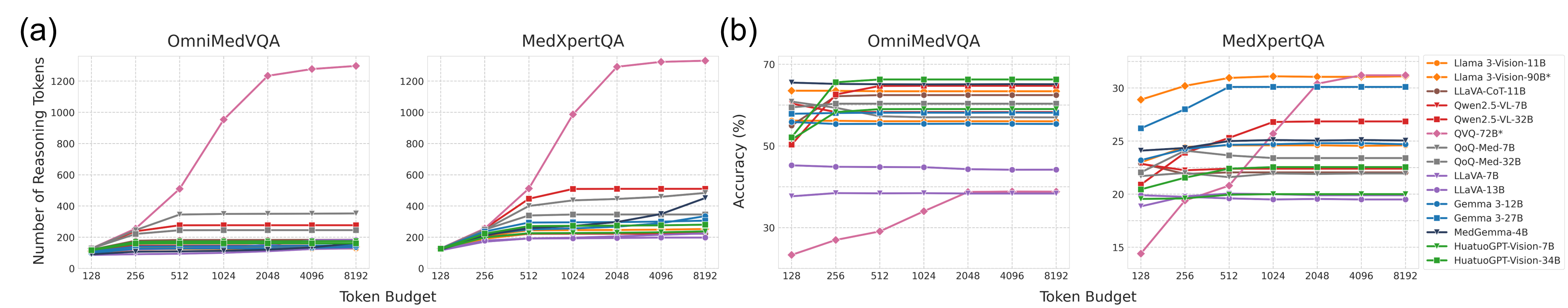}}
\caption{Test-time scaling of various VLMs across different token budgets on medical VQA benchmark datasets: (a) Average number of reasoning tokens used at different token budgets, (b) Accuracy of LLMs as a function of token budget.
$^*$ in the legend indicates models that are 4-bit quantized.}
\label{fig:vlm}
\end{figure*}

In contrast, m1 effectively utilizes a larger portion of the available token budget and exhibits substantial performance improvements as the token budget increases.
Notably, on the most challenging QA dataset in our study, MedXpertQA, m1-32B demonstrates clear test-time scalability, outperforming other models with even larger parameter counts as the token budget increases.
We attribute the difference between HuatuoGPT-o1 and m1 to differences in the nature of their respective training data.
Whle the reasoning traces used to train HuatuoGPT-o1 were generated by GPT-4o \cite{hurst2024gpt,chen2024huatuogpt}, those used for m1 were synthesized by DeepSeek-R1 \cite{guo2025deepseek,huang2025m1}.
These underlying models differ in their reasoning styles and output structures, likely leading to variations in the length and structure of CoT traces and influencing how each model utilizes tokens during inference.

\noindent{\textbf{General vs. Medical Models}}: As expected, medical LLMs generally outperform general domain LLMs on medical QA tasks.
However, in the case of MedCalc-Bench, the accuracy of medical models is generally lower than that of general domain models.
We hypothesize that this is because medical models are primarily fine-tuned on medical datasets, which often lack numerical reasoning tasks.
This fine-tuning process may impair the models' calculation abilities, leading to degraded performance on numerically intensive tasks such as MedCalc-Bench.
Nevertheless, performance can still be improved through test-time scaling with an increased token budget.

In summary, test-time scaling through increased token budgets can improve performance for certain reasoning LLMs.
However, many existing LLMs that benefit from this approach are not specifically designed for the medical domain.
This highlights the need to explore alternative test-time scaling strategies better suited for medical applications.

\subsubsection{Experiments wit VLMs}
Next, we investigate the test-time scaling in the medical domain using VLMs.
Similar to the previous experiment, we analyze the impact of increasing token budgets on model performance.

Figure~\ref{fig:vlm}(a) shows the average number of tokens used by VLMs during reasoning. Consistent with our findings from LLMs, most models, except QVQ, which is designed for reasoning, do not significantly increase their token usage, even when given a higher token budget.

Fig.~\ref{fig:vlm}(b) presents model accuracy on two datasets across different token budgets.
On OmniMedVQA, which is relatively simple and less demanding in reasoning, most models do not show any performance improvement as the token budget increases.
Only QVQ exhibits a consistent upward trend in accuracy with increased token budgets.
In contrast, on MedXpertQA, a dataset requiring more complex reasoning, we observe a more noticeable, though still limited, test-time scaling effect.
gain, only QVQ shows a clear benefit from increased token budgets, achieving the highest accuracy among the evaluated VLMs.

\begin{figure*}[!t]
\centerline{\includegraphics[width=\textwidth]{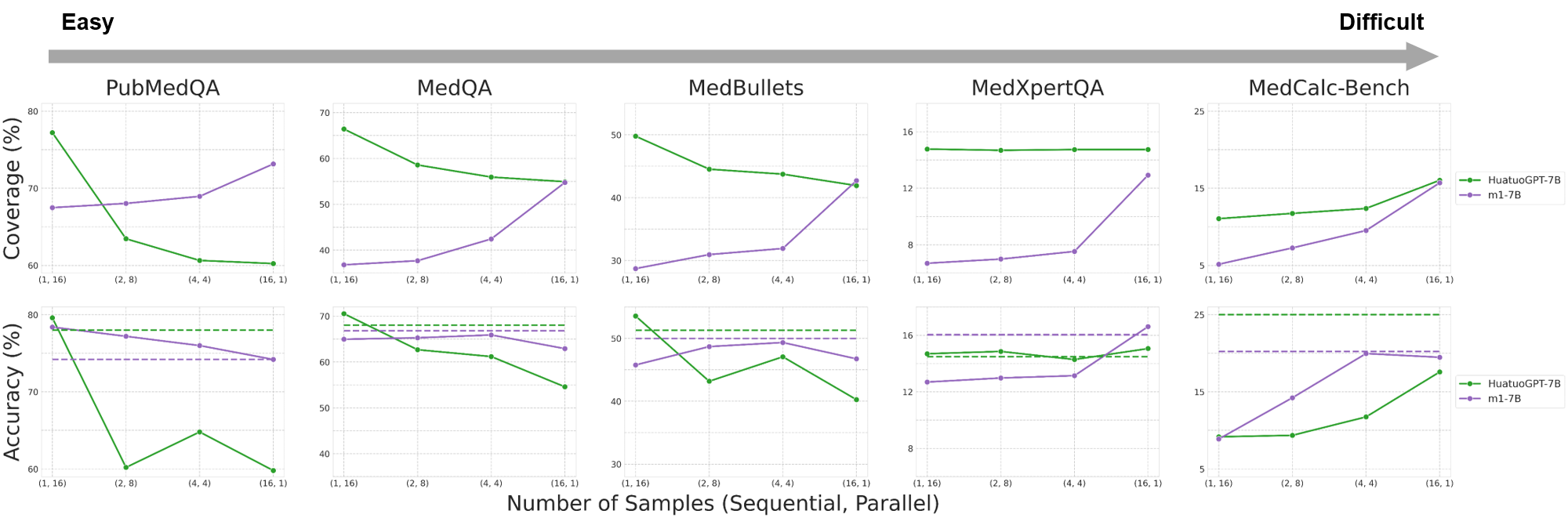}}
\caption{Comparison of sequential and parallel test-time scaling of LLMs on multiple medical benchmark datasets.
The dotted lines indicate the accuracy of each model when test-time scaling is applied by increasing the token budget to 8192.
The x-axis in each graph represents pairs indicating the number of iterative sequential samples and parallel samples, respectively.
Tasks are arranged from left to right in order of increasing difficulty and reasoning demands.}
\label{fig:llm_seq_para}
\end{figure*}

Interestingly, while QVQ displays test-time scaling on both benchmarks, its performance is paradoxically lower on the simpler OmniMedVQA and higher on the more challenging MedXpertQA.
Upon closer inspection of QVQ’s responses, we attribute this to a combination of model limitations and dataset characteristics.
Specifically, QVQ frequently fails to process the input images in OmniMedVQA correctly.
It sometimes outputs messages like \texttt{"I can't see the image"} (Supplementary Figure 3) or misinterprets prompts as part of the image content (e.g. \texttt{"There are also some text elements overlaid on the image, such as "You are a helpful assistant""}) (Supplementary Figures 4 and 5).
Additionally, due to these misinterpretations, QVQ often fails to choose from the provided answer options, instead responding with \texttt{"None of the above"} (Supplementary Figure 5), further reducing its accuracy.

In contrast, the performance of QVQ on MedXpertQA benefits from the presence of more informative textual clues within the questions, while the questions of OmniMedVQA are very simple (see Supplementary Figure 2).
These clues often enable the model to infer the correct answer even without fully utilizing the image, thus reducing the impact of its visual processing limitations.
As a result, QVQ achieves higher accuracy despite the greater task complexity.

Overall, our findings from the VLM experiments are consistent with those from the LLM experiments: test-time scaling by simply increasing the token budget provides limited benefits for most VLMs.
This approach shows effectiveness only for models explicitly designed for reasoning, such as QVQ, and even then, the performance gains in VLMs are considerably smaller than those observed in LLMs.
Notably, as in the case of LLMs, the effectiveness of test-time scaling in VLMs is more pronounced on more difficult tasks, while it has little to no impact on simpler tasks.

These results can be interpreted from two perspectives.
First, unlike LLMs, current VLMs may lack sufficiently developed reasoning capabilities, particularly in leveraging visual cues for complex decision-making.
While recent studies have proposed VLMs trained for reasoning, these models are often not explicitly trained to reason effectively using visual information, and thus struggle to interpret and analyze complex medical images as effectively as LLMs do with text.
Second, the limitations may stem from the current vision-language medical benchmarks, which may not be well-suited for evaluating reasoning ability.
For instance, OmniMedVQA primarily consists of relatively simple image-based questions that require minimal reasoning, thus failing to challenge the model’s reasoning capabilities.
Conversely, MedXpertQA presents highly challenging questions that often require the simultaneous analysis of multiple medical images.
Most contemporary VLMs still struggle with or entirely lack the ability to jointly and effectively reason over such complex visual cues in conjunction with textual information, which may explain relatively limited performance on this benchmark than that of LLM.
Future work should aim to develop stronger medical reasoning VLMs and more rigorous benchmarks for their evaluation.

\subsection{Sequential and Parallel Scaling}
\label{subsec:llm_seq_para}
In previous experiments, we observed that test-time scaling by simply increasing the token budget is ineffective for many models.
Except for certain reasoning models, most do not utilize the full available token budget during inference.
To address this limitation, we investigate alternative scaling strategies, iterative sequential scaling and parallel scaling, as described in Section~\ref{subsubsec:seq_para}.
In this experiment, we evaluate two medical reasoning LLMs: HuatuoGPT-o1-7B (denoted as \texttt{HuatuoGPT-7B}), and m1-7B-1k (denoted as \texttt{m1-7B}).

Fig.~\ref{fig:llm_seq_para} presents a comparison between sequential and parallel test-time scaling across multiple medical benchmarks.
On PubMedQA, the easiest QA task, both models show a decline in accuracy as the number of sequential sampling steps increases.
Moreover, the accuracy achieved with parallel scaling is higher than that obtained by simply increasing the token budget.
This suggests that additional iterations may not be beneficial for simpler tasks; in fact, excessive revisions can lead the models to deviate from initially correct answers.
In particular, for m1, accuracy decreases even though coverage increases with more sequential steps, indicating that while the model may identify the correct answer during sequential scaling, unnecessary revisions can lead to an incorrect final output.

For MedQA and MedBullets, HuatuoGPT-o1 maintains the same trend, showing limited improvement from either sequential scaling or increased token budgets.
In contrast, m1 achieves its highest accuracy when the token budget is evenly divided between sequential and parallel sampling (1:1 ratio), with performance comparable to that of simple token budget expansion.
These findings suggest that for tasks of moderate difficulty, sequential scaling may offer little advantage for models that naturally generate shorter outputs.
In comparison, models like m1, which rely on deeper reasoning, benefit more from a balanced approach that combines diverse generation with iterative refinement.

On the more challenging MedXpertQA benchmark, which demands complex reasoning, HuatuoGPT-o1 shows relatively consistent performance across all scaling strategies, including increased token budget, sequential scaling, and parallel scaling.
In contrast, m1 achieves its highest accuracy and coverage with fully sequential scaling, slightly surpassing its performance under the increased token budget condition.

On MedCalc-Bench, which also involves complex reasoning but with distinct characteristics, both models show improved performance as the number of sequential samples increases.
However, the highest accuracy is attained when applying test-time scaling via a larger token budget.
These findings suggest that step-by-step sequential refinement benefits reasoning-intensive question-answering tasks, while generating a single, extended reasoning path is more effective for tasks requiring precise calculations.

In summary, under constrained token budgets, parallel scaling proves more effective for models that inherently produce shorter reasoning traces, especially for simpler tasks.
In contrast, for models that rely on more extensive reasoning, sequential scaling or increasing the token budget becomes increasingly advantageous as task complexity grows.

\subsection{Impact of Test-Time Scaling on User-Driven Factors}
\begin{figure*}[!t]
\centerline{\includegraphics[width=\textwidth]{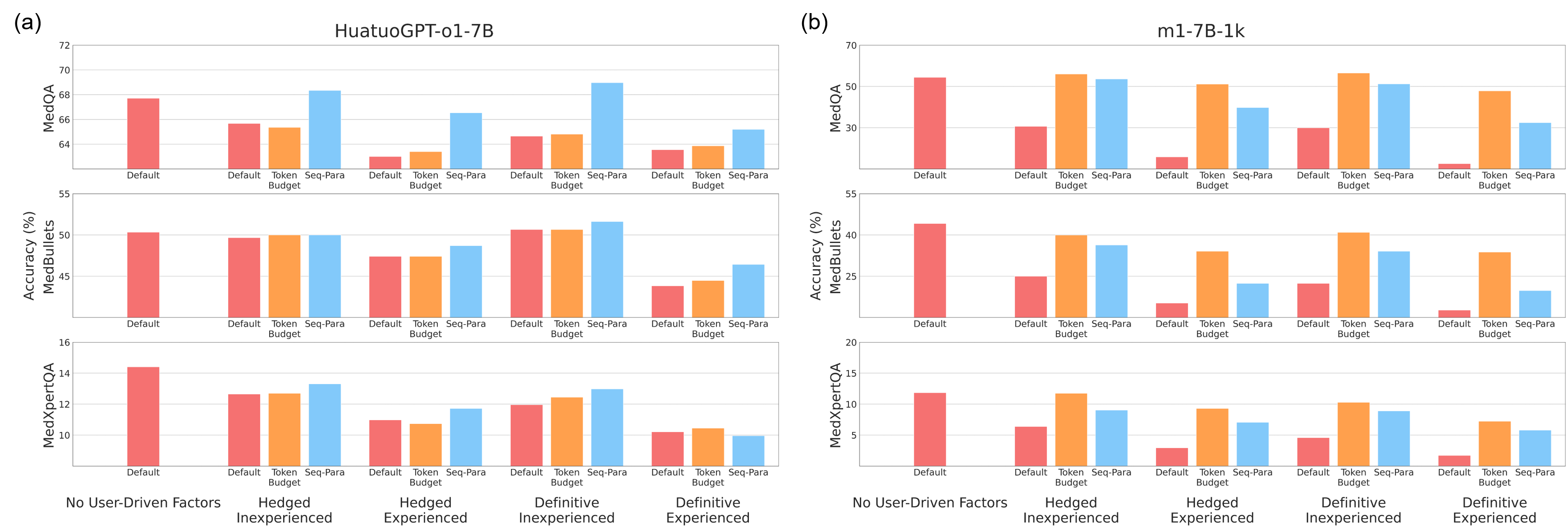}}
\caption{Test-time scaling under different user-driven factor types on medical QA benchmark datasets.
(a) HuatuoGPT-o1-7B and (b) m1-7B-1k.
The x-axis indicates the type of test-time scaling applied.
Default: No test-time scaling (maximum 512 tokens).
Token Budget: Increased token budget (maximum 8192 tokens).
Seq-Para: Combination of iterative sequential and parallel scaling using the optimal configuration described in Section~\ref{subsec:llm_seq_para} (maximum 512$\times$16 tokens).}
\label{fig:llm_perturb}
\end{figure*}

Finally, we investigate the impact of test-time scaling on robustness to user-driven factors.
Again, we evaluate two medical LLMs, HuatuoGPT-o1-7B and m1-7B-1k, across two types of test-time scaling strategies.
The first strategy simply increases the token budget by a factor of 16 from the default setting, as described in Section~\ref{subsec:llm_token_budget}.
The second strategy combines iterative sequential scaling and parallel scaling using the optimal configuration identified in Section~\ref{subsec:llm_seq_para}.
Fig.~\ref{fig:llm_perturb} shows the accuracy of both models under various user-driven perturbations and test-time scaling strategies.

First, Fig.~\ref{fig:llm_perturb}(a) shows that user-driven perturbation generally degrade the performance of HuatuoGPT-o1 when no test-time scaling strategy is applied.
When test-time scaling is applied by simply increasing the token budget, the improvement is minimal, and in some cases, performance even declines.
In contrast, applying the optimal test-time scaling strategy significantly improves robustness against user-driven factors in most scenarios.
The benefit is particularly pronounced when the task is easier, such as in MedQA.
This is likely because HuatuoGPT-o1 tends to generate relatively short responses by default, and merely increasing the token budget does not effectively encourage the model to utilize the additional tokens for deeper reasoning.

On the other hand, m1 also exhibits degraded performance in the presence of user-driven factors.
However, unlike HuatuoGPT-o1, increasing the token budget proves more effective than applying the optimal sequential-parallel scaling strategy for the m1 model.
This is due to m1’s inherent tendency to generate long CoT reasoning when solving problems.
Since sequential-parallel scaling produces shorter responses at each iteration, it may hinder m1 from fully developing its reasoning and reaching a well-considered conclusion.

In summary, model behavior varies depending on the chosen test-time scaling strategy when user-driven factors exist.
This highlights the importance of selecting an appropriate test-time scaling method tailored to each model type to enhance robustness against such perturbations.
Nonetheless, some common patterns emerge across models.
For relatively easy tasks (e.g., MedQA), both models can recover performance comparable to that without user-driven interference through test-time scaling.
In contrast, for more challenging tasks (e.g., MedXpertQA), the performance gap between conditions with and without user-driven factors remains substantial, even when test-time scaling is applied.
Notably, models are particularly sensitive to the expertise level of the physician providing the additional input.
This suggests that as task complexity increases, LLMs may struggle to initiate reasoning with high confidence, leading them to heavily rely on the initial context or cues provided by the user.
Consequently, when such cues are misleading, the model is more likely to follow an incorrect reasoning path and, due to the increased complexity, has greater difficulty recovering and arriving at the correct answer.
These findings emphasize the amplified impact of user-driven perturbations in difficult scenarios and highlight the importance of minimizing such perturbations by providing only neutral and essential information when prompting LLMs for complex or high-stakes medical question answering tasks.

\section{Conclusion}
\label{sec:conclusion}
In this paper, we conducted a comprehensive investigation of test-time scaling in the medical domain across various types of LLMs and VLMs.
Our experiments demonstrate that longer reasoning is not universally beneficial across all medical tasks.
For simpler tasks, extensive reasoning is unnecessary, and models that produce concise reasoning traces offer better computational efficiency.
In contrast, for complex and challenging tasks, generating longer CoT reasoning leads to better performance, with models capable of producing extended reasoning traces outperforming their counterparts.
Additionally, we observed that medical LLMs tend to perform worse on medical calculation tasks compared to general domain LLMs, likely due to their fine-tuning process, which often focuses on textual medical knowledge and may diminish numerical reasoning capabilities.

Furthermore, we examined whether sequential or parallel test-time scaling provides greater benefits.
Our findings indicate that the optimal test-time scaling strategy varies depending on both the model type and the difficulty of the task.
Additionally, we found that selecting an appropriate scaling strategy is crucial for enhancing model robustness against user-driven factors, such as misleading or biased information in the prompts.
Notably, the impact of these user-driven factors becomes more pronounced as task difficulty increases, emphasizing the importance of minimizing perturbations by providing only neutral and essential information for complex tasks.
A summary of our findings is presented in TABLE~\ref{tab:guide_task} and TABLE~\ref{tab:guide_model}.

Although we conducted a broad investigation of test-time scaling in the medical domain, several limitations remain.
First, our study focused on scaling strategies such as increasing the token budget, iterative self-revision via sequential scaling, and generating multiple responses in parallel.
However, other promising strategies, such as employing trained verifiers to select the most accurate response among candidates, were not explored.
In addition, compared to LLMs, VLMs require more comprehensive evaluation under test-time scaling.
While our results show that test-time scaling can improve robustness to user-driven factors, it was not sufficient to fully restore performance to baseline levels observed without such perturbations.
Future work should investigate more advanced or hybrid test-time scaling techniques that can further mitigate the impact of user-driven factors.
Finally, our experiments were limited to open-source LLMs, while proprietary models such as GPT, Claude, or Gemini were not included.
Extending test-time scaling evaluations to these models represents an important direction for future research in the medical domain.

Despite these limitations, we believe our study offers valuable insights and practical guidance for applying test-time scaling in the medical domain.

\begin{table}[!t]
    \centering
    \caption{Recommended test-time scaling strategies based on task difficulty in the medical domain.}
    \label{tab:guide_task}
    \resizebox{0.99\columnwidth}{!}{
    \begin{tabular}{c|c}
    \hline
    Difficulty  & Recommendation    \\
    \hline\hline
    Easy    & \makecell[l]{$\bullet$ Prefer parallel scaling over simply increasing the token budget\\or using iterative sequential scaling\\\\$\bullet$ Avoid user-driven factors that use a definitive tone and reflect\\expert-level input}\\
    \hline
    Intermediate    & \makecell[l]{$\bullet$ Adapt the scaling strategy based on the model's token usage \\pattern, combining sequential and parallel scaling as appropriate\\\\$\bullet$ Avoid user-driven factors that use a definitive tone}\\
    \hline
    Difficult   & \makecell[l]{$\bullet$ Apply iterative sequential scaling for question answering tasks\\to enhance step-by-step reasoning\\\\$\bullet$ Increase the token budget for calculation tasks that require\\extended reasoning\\\\$\bullet$ Avoid all types of user-driven factors and provide only neutral\\and essential information}\\
    \hline
    \end{tabular}
    }
\end{table}

\begin{table*}[!t]
    \centering
    \caption{Recommended test-time scaling strategies based on model types in the medical domain.}
    \label{tab:guide_model}
    \resizebox{0.99\textwidth}{!}{
    \begin{tabular}{c|c|c|c}
    \hline
    Model Type  & Strength  &  Weakness  & Recommendation    \\
    \hline\hline
    General, Non-Reasoning    & \makecell[l]{$\bullet$ Performs well on medical calculation tasks}  & \makecell[l]{$\bullet$ Limited medical knowledge\\\\$\bullet$ Relatively low performance on reasoning-intensive\\medical question answering tasks\\\\$\bullet$ Utilize a small number of tokens for reasoning}  & \makecell[l]{$\bullet$ Use primarily for medical calculation tasks\\\\$\bullet$ Prefer parallel scaling over other strategies}\\
    \hline
    General, Reasoning    & \makecell[l]{$\bullet$ Performs well on medical calculation tasks\\\\$\bullet$ High performance on reasoning-intensive tasks}  & \makecell[l]{$\bullet$ Limited medical knowledge\\\\$\bullet$ Relatively low performance on reasoning-intensive\\medical question answering tasks\\\\$\bullet$ Underperforms when using small model sizes}  & \makecell[l]{$\bullet$ Use primarily for medical calculation tasks\\\\$\bullet$ Adjust scaling strategy based on the model’s token\\usage pattern and task difficulty, combining sequential\\and parallel scaling as needed\\\\$\bullet$ Use sufficiently large models (over 30B parameters)}\\
    \hline
    Medical, Non-Reasoning    & \makecell[l]{$\bullet$ Performs well on medical question-answering tasks}  & \makecell[l]{$\bullet$ Relatively low performance on reasoning-intensive\\medical calculation tasks\\\\$\bullet$ Utilize a small number of tokens for reasoning}  & \makecell[l]{$\bullet$ Use primarily for medical question answering tasks\\\\$\bullet$ Prefer parallel scaling over other strategies}\\
    \hline
    Medical, Reasoning    & \makecell[l]{$\bullet$ Performs well on medical question-answering tasks\\\\$\bullet$ High performance on reasoning-intensive tasks}  & \makecell[l]{$\bullet$ Relatively low performance on reasoning-intensive\\medical calculation tasks}  & \makecell[l]{$\bullet$ Use primarily for medical question answering tasks\\\\$\bullet$ Adjust scaling strategy based on the model’s token\\usage pattern and task difficulty, combining sequential\\and parallel scaling as needed}\\
    \hline
    \end{tabular}
    }
\end{table*}

\section*{Acknowledgment}
We thank Kyung Ho Lim for the valuable advice on the experiments involving user-driven factors.

\bibliographystyle{IEEEtran}
\bibliography{ref}

\begin{thebibliography}{10}
\providecommand{\url}[1]{#1}
\csname url@samestyle\endcsname
\providecommand{\newblock}{\relax}
\providecommand{\bibinfo}[2]{#2}
\providecommand{\BIBentrySTDinterwordspacing}{\spaceskip=0pt\relax}
\providecommand{\BIBentryALTinterwordstretchfactor}{4}
\providecommand{\BIBentryALTinterwordspacing}{\spaceskip=\fontdimen2\font plus
\BIBentryALTinterwordstretchfactor\fontdimen3\font minus \fontdimen4\font\relax}
\providecommand{\BIBforeignlanguage}[2]{{%
\expandafter\ifx\csname l@#1\endcsname\relax
\typeout{** WARNING: IEEEtran.bst: No hyphenation pattern has been}%
\typeout{** loaded for the language `#1'. Using the pattern for}%
\typeout{** the default language instead.}%
\else
\language=\csname l@#1\endcsname
\fi
#2}}
\providecommand{\BIBdecl}{\relax}
\BIBdecl

\bibitem{brown2020language}
T.~Brown, B.~Mann, N.~Ryder, M.~Subbiah, J.~D. Kaplan, P.~Dhariwal, A.~Neelakantan, P.~Shyam, G.~Sastry, A.~Askell \emph{et~al.}, ``Language models are few-shot learners,'' \emph{Advances in neural information processing systems}, vol.~33, pp. 1877--1901, 2020.

\bibitem{achiam2023gpt}
J.~Achiam, S.~Adler, S.~Agarwal, L.~Ahmad, I.~Akkaya, F.~L. Aleman, D.~Almeida, J.~Altenschmidt, S.~Altman, S.~Anadkat \emph{et~al.}, ``{GPT}-4 technical report,'' \emph{arXiv preprint arXiv:2303.08774}, 2023.

\bibitem{chiang2023vicuna}
W.-L. Chiang, Z.~Li, Z.~Lin, Y.~Sheng, Z.~Wu, H.~Zhang, L.~Zheng, S.~Zhuang, Y.~Zhuang, J.~E. Gonzalez \emph{et~al.}, ``Vicuna: An open-source chatbot impressing gpt-4 with 90\%* chatgpt quality,'' \emph{See https://vicuna. lmsys. org (accessed 14 April 2023)}, vol.~2, no.~3, p.~6, 2023.

\bibitem{touvron2023llama}
H.~Touvron, L.~Martin, K.~Stone, P.~Albert, A.~Almahairi, Y.~Babaei, N.~Bashlykov, S.~Batra, P.~Bhargava, S.~Bhosale \emph{et~al.}, ``Llama 2: Open foundation and fine-tuned chat models,'' \emph{arXiv preprint arXiv:2307.09288}, 2023.

\bibitem{hurst2024gpt}
A.~Hurst, A.~Lerer, A.~P. Goucher, A.~Perelman, A.~Ramesh, A.~Clark, A.~Ostrow, A.~Welihinda, A.~Hayes, A.~Radford \emph{et~al.}, ``{GPT}-4o system card,'' \emph{arXiv preprint arXiv:2410.21276}, 2024.

\bibitem{jaech2024openai}
A.~Jaech, A.~Kalai, A.~Lerer, A.~Richardson, A.~El-Kishky, A.~Low, A.~Helyar, A.~Madry, A.~Beutel, A.~Carney \emph{et~al.}, ``{OpenAI} o1 system card,'' \emph{arXiv preprint arXiv:2412.16720}, 2024.

\bibitem{grattafiori2024llama}
A.~Grattafiori, A.~Dubey, A.~Jauhri, A.~Pandey, A.~Kadian, A.~Al-Dahle, A.~Letman, A.~Mathur, A.~Schelten, A.~Vaughan \emph{et~al.}, ``The {Llama} 3 herd of models,'' \emph{arXiv e-prints}, pp. arXiv--2407, 2024.

\bibitem{guo2025deepseek}
D.~Guo, D.~Yang, H.~Zhang, J.~Song, R.~Zhang, R.~Xu, Q.~Zhu, S.~Ma, P.~Wang, X.~Bi \emph{et~al.}, ``{DeepSeek-R1}: {Incentivizing} reasoning capability in {LLMs} via reinforcement learning,'' \emph{arXiv preprint arXiv:2501.12948}, 2025.

\bibitem{yang2025qwen25}
A.~Yang, B.~Yang, B.~Zhang, B.~Hui, B.~Zheng, B.~Yu, C.~Li, D.~Liu, F.~Huang, H.~Wei, H.~Lin, J.~Yang, J.~Tu, J.~Zhang, J.~Yang, J.~Yang, J.~Zhou, J.~Lin, K.~Dang, K.~Lu, K.~Bao, K.~Yang, L.~Yu, M.~Li, M.~Xue, P.~Zhang, Q.~Zhu, R.~Men, R.~Lin, T.~Li, T.~Tang, T.~Xia, X.~Ren, X.~Ren, Y.~Fan, Y.~Su, Y.~Zhang, Y.~Wan, Y.~Liu, Z.~Cui, Z.~Zhang, and Z.~Qiu, ``Qwen2.5 technical report,'' \emph{arXiv preprint arXiv:2412.15115}, 2025.

\bibitem{alayrac2022flamingo}
J.-B. Alayrac, J.~Donahue, P.~Luc, A.~Miech, I.~Barr, Y.~Hasson, K.~Lenc, A.~Mensch, K.~Millican, M.~Reynolds \emph{et~al.}, ``Flamingo: {A} visual language model for few-shot learning,'' \emph{Advances in neural information processing systems}, vol.~35, pp. 23\,716--23\,736, 2022.

\bibitem{li2022blip}
J.~Li, D.~Li, C.~Xiong, and S.~Hoi, ``{BLIP}: {Bootstrapping} language-image pre-training for unified vision-language understanding and generation,'' in \emph{International conference on machine learning}.\hskip 1em plus 0.5em minus 0.4em\relax PMLR, 2022, pp. 12\,888--12\,900.

\bibitem{li2023blip}
J.~Li, D.~Li, S.~Savarese, and S.~Hoi, ``{BLIP}-2: {Bootstrapping} language-image pre-training with frozen image encoders and large language models,'' in \emph{International conference on machine learning}.\hskip 1em plus 0.5em minus 0.4em\relax PMLR, 2023, pp. 19\,730--19\,742.

\bibitem{liu2023visual}
H.~Liu, C.~Li, Q.~Wu, and Y.~J. Lee, ``Visual instruction tuning,'' \emph{Advances in neural information processing systems}, vol.~36, pp. 34\,892--34\,916, 2023.

\bibitem{team2024gemini}
G.~Team, P.~Georgiev, V.~I. Lei, R.~Burnell, L.~Bai, A.~Gulati, G.~Tanzer, D.~Vincent, Z.~Pan, S.~Wang \emph{et~al.}, ``Gemini 1.5: Unlocking multimodal understanding across millions of tokens of context,'' \emph{arXiv preprint arXiv:2403.05530}, 2024.

\bibitem{wang2024qwen2}
P.~Wang, S.~Bai, S.~Tan, S.~Wang, Z.~Fan, J.~Bai, K.~Chen, X.~Liu, J.~Wang, W.~Ge \emph{et~al.}, ``{Qwen2-VL}: {Enhancing} vision-language model's perception of the world at any resolution,'' \emph{arXiv preprint arXiv:2409.12191}, 2024.

\bibitem{xue2024xgen}
L.~Xue, M.~Shu, A.~Awadalla, J.~Wang, A.~Yan, S.~Purushwalkam, H.~Zhou, V.~Prabhu, Y.~Dai, M.~S. Ryoo \emph{et~al.}, ``{xGen-MM} ({BLIP}-3): {A} family of open large multimodal models,'' \emph{arXiv preprint arXiv:2408.08872}, 2024.

\bibitem{muennighoff2025s1}
N.~Muennighoff, Z.~Yang, W.~Shi, X.~L. Li, L.~Fei-Fei, H.~Hajishirzi, L.~Zettlemoyer, P.~Liang, E.~Cand{\`e}s, and T.~Hashimoto, ``s1: {Simple} test-time scaling,'' \emph{arXiv preprint arXiv:2501.19393}, 2025.

\bibitem{snell2025scaling}
\BIBentryALTinterwordspacing
C.~V. Snell, J.~Lee, K.~Xu, and A.~Kumar, ``Scaling {LLM} test-time compute optimally can be more effective than scaling parameters for reasoning,'' in \emph{The Thirteenth International Conference on Learning Representations}, 2025. [Online]. Available: \url{https://openreview.net/forum?id=4FWAwZtd2n}
\BIBentrySTDinterwordspacing

\bibitem{yang2502towards}
W.~Yang, S.~Ma, Y.~Lin, and F.~Wei, ``Towards thinking-optimal scaling of test-time compute for {LLM} reasoning,'' \emph{URL https://arxiv. org/abs/2502.18080}, 2025.

\bibitem{zeng2025revisiting}
Z.~Zeng, Q.~Cheng, Z.~Yin, Y.~Zhou, and X.~Qiu, ``Revisiting the test-time scaling of o1-like models: {Do} they truly possess test-time scaling capabilities?'' \emph{arXiv preprint arXiv:2502.12215}, 2025.

\bibitem{schulman2017proximal}
J.~Schulman, F.~Wolski, P.~Dhariwal, A.~Radford, and O.~Klimov, ``Proximal policy optimization algorithms,'' \emph{arXiv preprint arXiv:1707.06347}, 2017.

\bibitem{rafailov2023direct}
R.~Rafailov, A.~Sharma, E.~Mitchell, C.~D. Manning, S.~Ermon, and C.~Finn, ``Direct preference optimization: {Your} language model is secretly a reward model,'' \emph{Advances in Neural Information Processing Systems}, vol.~36, pp. 53\,728--53\,741, 2023.

\bibitem{shao2024deepseekmath}
Z.~Shao, P.~Wang, Q.~Zhu, R.~Xu, J.~Song, X.~Bi, H.~Zhang, M.~Zhang, Y.~Li, Y.~Wu \emph{et~al.}, ``{DeepSeekMath}: {Pushing} the limits of mathematical reasoning in open language models,'' \emph{arXiv preprint arXiv:2402.03300}, 2024.

\bibitem{moon2022multi}
J.~H. Moon, H.~Lee, W.~Shin, Y.-H. Kim, and E.~Choi, ``Multi-modal understanding and generation for medical images and text via vision-language pre-training,'' \emph{IEEE Journal of Biomedical and Health Informatics}, vol.~26, no.~12, pp. 6070--6080, 2022.

\bibitem{li2023llava}
C.~Li, C.~Wong, S.~Zhang, N.~Usuyama, H.~Liu, J.~Yang, T.~Naumann, H.~Poon, and J.~Gao, ``{LLaVA-Med}: {Training} a large language-and-vision assistant for biomedicine in one day,'' \emph{Advances in Neural Information Processing Systems}, vol.~36, pp. 28\,541--28\,564, 2023.

\bibitem{park2024self}
S.~Park, E.~S. Lee, K.~S. Shin, J.~E. Lee, and J.~C. Ye, ``Self-supervised multi-modal training from uncurated images and reports enables monitoring {AI} in radiology,'' \emph{Medical Image Analysis}, vol.~91, p. 103021, 2024.

\bibitem{chen2024huatuogpt}
J.~Chen, Z.~Cai, K.~Ji, X.~Wang, W.~Liu, R.~Wang, J.~Hou, and B.~Wang, ``{HuatuoGPT}-o1, towards medical complex reasoning with {LLMs},'' \emph{arXiv preprint arXiv:2412.18925}, 2024.

\bibitem{zhang2024generalist}
K.~Zhang, R.~Zhou, E.~Adhikarla, Z.~Yan, Y.~Liu, J.~Yu, Z.~Liu, X.~Chen, B.~D. Davison, H.~Ren \emph{et~al.}, ``A generalist vision--language foundation model for diverse biomedical tasks,'' \emph{Nature Medicine}, pp. 1--13, 2024.

\bibitem{zhang2024ultramedical}
K.~Zhang, S.~Zeng, E.~Hua, N.~Ding, Z.-R. Chen, Z.~Ma, H.~Li, G.~Cui, B.~Qi, X.~Zhu \emph{et~al.}, ``{UltraMedical}: {Building} specialized generalists in biomedicine,'' \emph{Advances in Neural Information Processing Systems}, vol.~37, pp. 26\,045--26\,081, 2024.

\bibitem{jiang2025meds}
S.~Jiang, Y.~Liao, Z.~Chen, Y.~Zhang, Y.~Wang, and Y.~Wang, ``Meds$^{3}$: {Towards} medical small language models with self-evolved slow thinking,'' \emph{arXiv preprint arXiv:2501.12051}, 2025.

\bibitem{dai2025qoq}
W.~Dai, P.~Chen, C.~Ekbote, and P.~P. Liang, ``{QoQ-Med}: {Building} multimodal clinical foundation models with domain-aware {GRPO} training,'' \emph{arXiv preprint arXiv:2506.00711}, 2025.

\bibitem{lai2025med}
Y.~Lai, J.~Zhong, M.~Li, S.~Zhao, and X.~Yang, ``{Med-R1}: {Reinforcement} learning for generalizable medical reasoning in vision-language models,'' \emph{arXiv preprint arXiv:2503.13939}, 2025.

\bibitem{pan2025medvlm}
J.~Pan, C.~Liu, J.~Wu, F.~Liu, J.~Zhu, H.~B. Li, C.~Chen, C.~Ouyang, and D.~Rueckert, ``{MedVLM-R1}: {Incentivizing} medical reasoning capability of vision-language models ({VLMs}) via reinforcement learning,'' \emph{arXiv preprint arXiv:2502.19634}, 2025.

\bibitem{huang2025m1}
X.~Huang, J.~Wu, H.~Liu, X.~Tang, and Y.~Zhou, ``m1: {Unleash} the potential of test-time scaling for medical reasoning with large language models,'' \emph{arXiv preprint arXiv:2504.00869}, 2025.

\bibitem{huang2025o1}
Z.~Huang, G.~Geng, S.~Hua, Z.~Huang, H.~Zou, S.~Zhang, P.~Liu, and X.~Zhang, ``O1 replication journey--{Part} 3: {Inference}-time scaling for medical reasoning,'' \emph{arXiv preprint arXiv:2501.06458}, 2025.

\bibitem{balachandran2025inference}
V.~Balachandran, J.~Chen, L.~Chen, S.~Garg, N.~Joshi, Y.~Lara, J.~Langford, B.~Nushi, V.~Vineet, Y.~Wu \emph{et~al.}, ``Inference-time scaling for complex tasks: {Where} we stand and what lies ahead,'' \emph{arXiv preprint arXiv:2504.00294}, 2025.

\bibitem{chen2024huatuogptv}
J.~Chen, C.~Gui, R.~Ouyang, A.~Gao, S.~Chen, G.~H. Chen, X.~Wang, R.~Zhang, Z.~Cai, K.~Ji \emph{et~al.}, ``{HuatuoGPT-Vision}, towards injecting medical visual knowledge into multimodal {LLMs} at scale,'' \emph{arXiv preprint arXiv:2406.19280}, 2024.

\bibitem{medgemma}
\BIBentryALTinterwordspacing
G.~Health, ``{MedGemma}: {Advanced} {AI} models for medical text and image analysis,'' May 2025. [Online]. Available: \url{https://medgemma.org/}
\BIBentrySTDinterwordspacing

\bibitem{bai2025qwen2}
S.~Bai, K.~Chen, X.~Liu, J.~Wang, W.~Ge, S.~Song, K.~Dang, P.~Wang, S.~Wang, J.~Tang \emph{et~al.}, ``{Qwen2.5-VL} technical report,'' \emph{arXiv preprint arXiv:2502.13923}, 2025.

\bibitem{team2025gemma}
G.~Team, A.~Kamath, J.~Ferret, S.~Pathak, N.~Vieillard, R.~Merhej, S.~Perrin, T.~Matejovicova, A.~Ram{\'e}, M.~Rivi{\`e}re \emph{et~al.}, ``Gemma 3 technical report,'' \emph{arXiv preprint arXiv:2503.19786}, 2025.

\bibitem{xu2024llava}
G.~Xu, P.~Jin, L.~Hao, Y.~Song, L.~Sun, and L.~Yuan, ``{LLaVA-o1}: {Let} vision language models reason step-by-step,'' \emph{arXiv preprint arXiv:2411.10440}, 2024.

\bibitem{qvq-72b-preview}
\BIBentryALTinterwordspacing
Q.~Team, ``{QVQ}: {To} see the world with wisdom,'' December 2024. [Online]. Available: \url{https://qwenlm.github.io/blog/qvq-72b-preview/}
\BIBentrySTDinterwordspacing

\bibitem{young2024yi}
A.~Young, B.~Chen, C.~Li, C.~Huang, G.~Zhang, G.~Zhang, G.~Wang, H.~Li, J.~Zhu, J.~Chen \emph{et~al.}, ``Yi: {Open} foundation models by 01. ai,'' \emph{arXiv preprint arXiv:2403.04652}, 2024.

\bibitem{dai2025climb}
W.~Dai, P.~Chen, M.~Lu, D.~Li, H.~Wei, H.~Cui, and P.~P. Liang, ``{CLIMB}: {Data} foundations for large scale multimodal clinical foundation models,'' \emph{arXiv preprint arXiv:2503.07667}, 2025.

\bibitem{jin2019pubmedqa}
Q.~Jin, B.~Dhingra, Z.~Liu, W.~W. Cohen, and X.~Lu, ``{PubMedQA}: {A} dataset for biomedical research question answering,'' \emph{arXiv preprint arXiv:1909.06146}, 2019.

\bibitem{jin2021disease}
D.~Jin, E.~Pan, N.~Oufattole, W.-H. Weng, H.~Fang, and P.~Szolovits, ``What disease does this patient have? {A} large-scale open domain question answering dataset from medical exams,'' \emph{Applied Sciences}, vol.~11, no.~14, p. 6421, 2021.

\bibitem{chen2025benchmarking}
H.~Chen, Z.~Fang, Y.~Singla, and M.~Dredze, ``Benchmarking large language models on answering and explaining challenging medical questions,'' in \emph{Proceedings of the 2025 Conference of the Nations of the Americas Chapter of the Association for Computational Linguistics: Human Language Technologies (Volume 1: Long Papers)}, 2025, pp. 3563--3599.

\bibitem{zuo2025medxpertqa}
Y.~Zuo, S.~Qu, Y.~Li, Z.~Chen, X.~Zhu, E.~Hua, K.~Zhang, N.~Ding, and B.~Zhou, ``{MedXpertQA}: {Benchmarking} expert-level medical reasoning and understanding,'' \emph{arXiv preprint arXiv:2501.18362}, 2025.

\bibitem{khandekar2024medcalc}
N.~Khandekar, Q.~Jin, G.~Xiong, S.~Dunn, S.~Applebaum, Z.~Anwar, M.~Sarfo-Gyamfi, C.~Safranek, A.~Anwar, A.~Zhang \emph{et~al.}, ``{MedCalc-Bench}: {Evaluating} large language models for medical calculations,'' \emph{Advances in Neural Information Processing Systems}, vol.~37, pp. 84\,730--84\,745, 2024.

\bibitem{lim2025susceptibility}
K.~H. Lim, U.~Kang, X.~Li, J.~S. Kim, Y.-C. Jung, S.~Park, and B.-H. Kim, ``Susceptibility of large language models to user-driven factors in medical queries,'' \emph{arXiv preprint arXiv:2503.22746}, 2025.

\bibitem{hu2024omnimedvqa}
Y.~Hu, T.~Li, Q.~Lu, W.~Shao, J.~He, Y.~Qiao, and P.~Luo, ``{OmniMedVQA}: A new large-scale comprehensive evaluation benchmark for medical {LVLM},'' in \emph{Proceedings of the IEEE/CVF Conference on Computer Vision and Pattern Recognition}, 2024, pp. 22\,170--22\,183.

\bibitem{ouyang2022training}
L.~Ouyang, J.~Wu, X.~Jiang, D.~Almeida, C.~Wainwright, P.~Mishkin, C.~Zhang, S.~Agarwal, K.~Slama, A.~Ray \emph{et~al.}, ``Training language models to follow instructions with human feedback,'' \emph{Advances in neural information processing systems}, vol.~35, pp. 27\,730--27\,744, 2022.

\bibitem{ethayarajh2024kto}
K.~Ethayarajh, W.~Xu, N.~Muennighoff, D.~Jurafsky, and D.~Kiela, ``{KTO}: {Model} alignment as prospect theoretic optimization,'' \emph{arXiv preprint arXiv:2402.01306}, 2024.

\bibitem{chen2025reasoning}
Y.~Chen, J.~Benton, A.~Radhakrishnan, J.~Uesato, C.~Denison, J.~Schulman, A.~Somani, P.~Hase, M.~Wagner, F.~Roger \emph{et~al.}, ``Reasoning models don't always say what they think,'' \emph{arXiv preprint arXiv:2505.05410}, 2025.

\bibitem{thapa2025disentangling}
R.~Thapa, Q.~Wu, K.~Wu, H.~Zhang, A.~Zhang, E.~Wu, H.~Ye, S.~Bedi, N.~Aresh, J.~Boen \emph{et~al.}, ``Disentangling reasoning and knowledge in medical large language models,'' \emph{arXiv preprint arXiv:2505.11462}, 2025.

\end{thebibliography}

\setcounter{figure}{0}
\renewcommand{\figurename}{Supplementary Figure}
\renewcommand{\thefigure}{\arabic{figure}}

\begin{figure*}[!ht]
    \centering
    \includegraphics[width=0.9\linewidth]{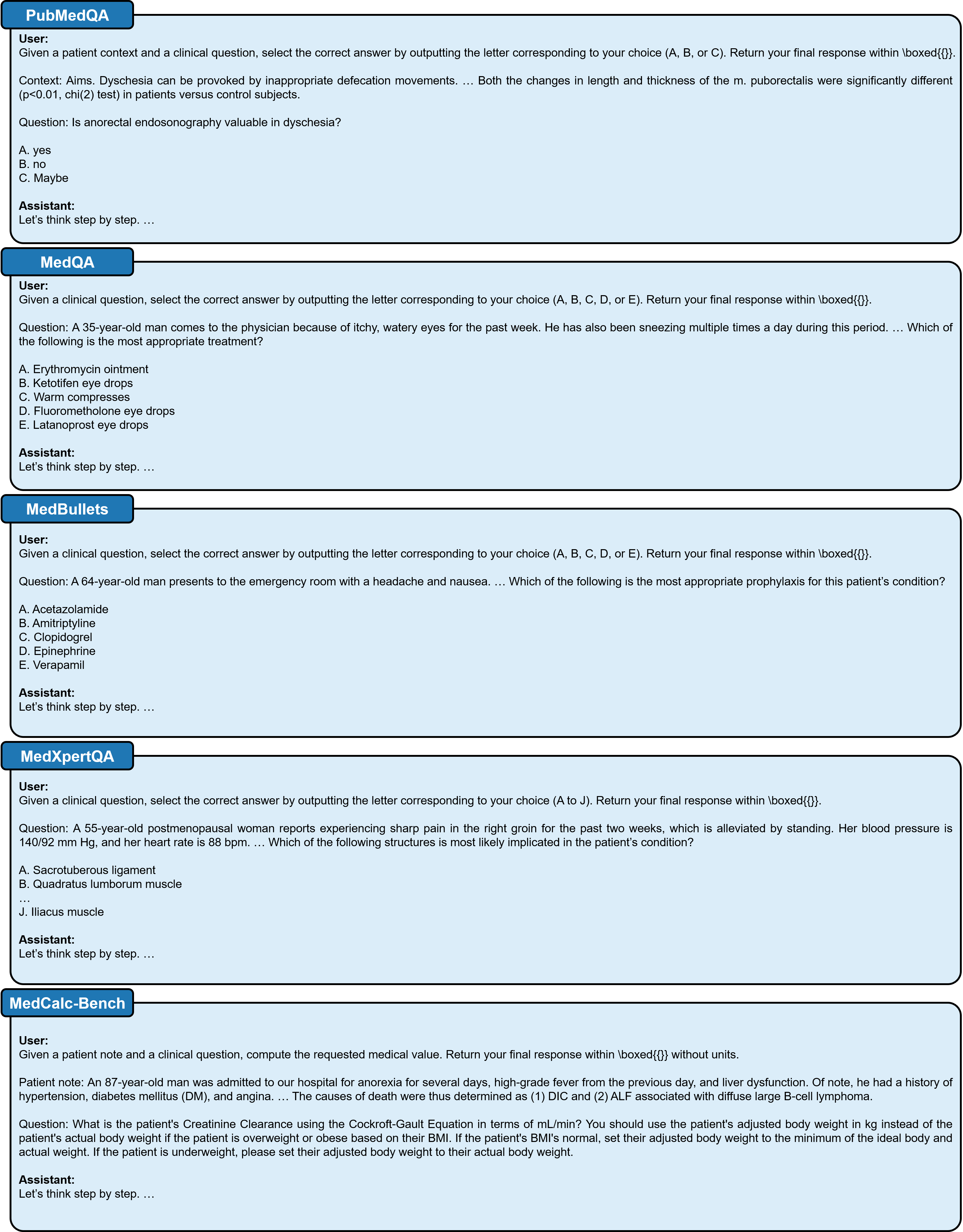}
    \caption{Prompts for text-only medical benchmark datasets.}
    \label{fig:prompt_llm}
\end{figure*}

\begin{figure*}[!ht]
    \centering
    \includegraphics[width=0.9\linewidth]{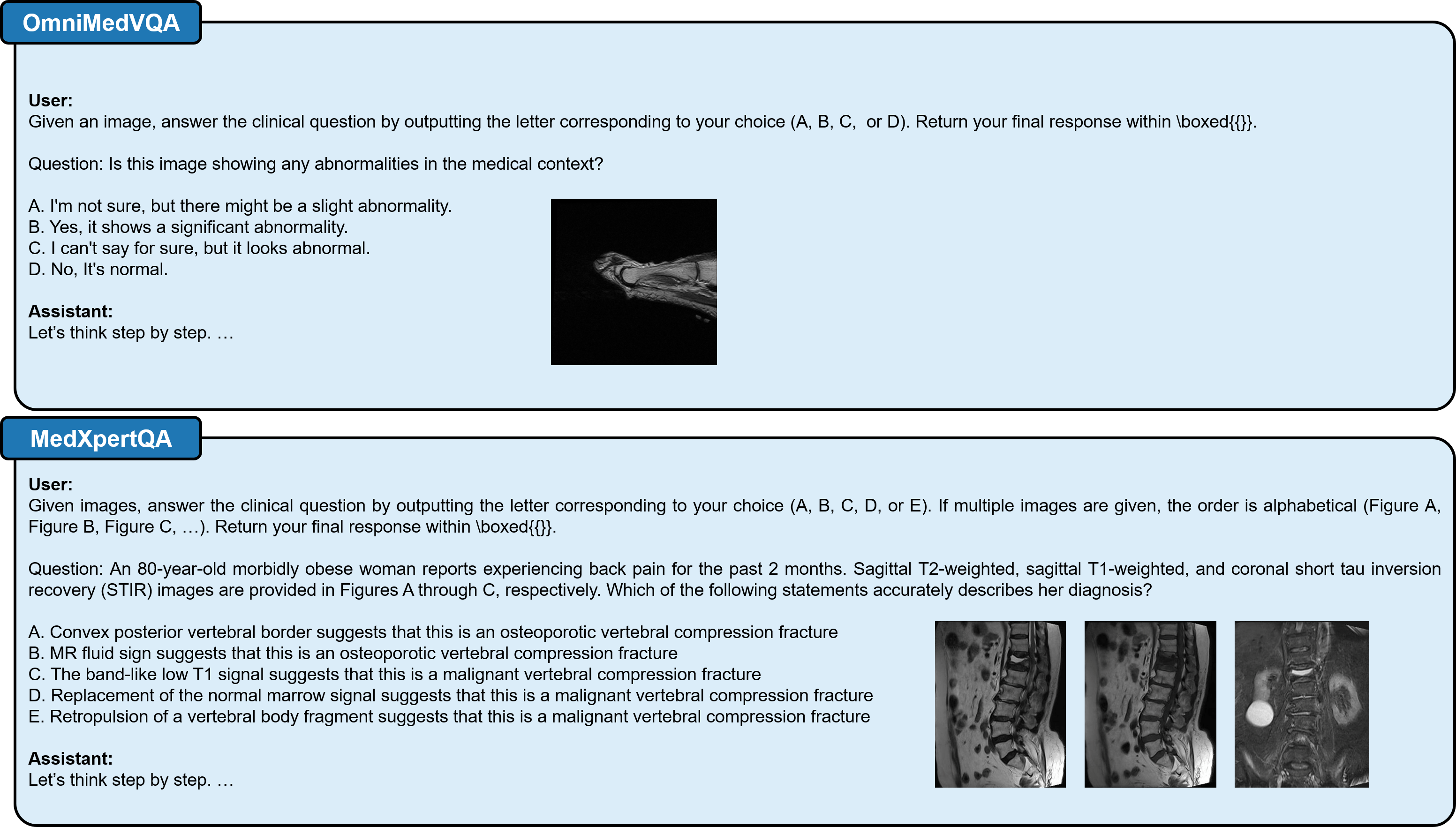}
    \caption{Prompts for vision-text medical benchmark datasets.}
    \label{fig:prompt_vlm}
\end{figure*}

\begin{figure*}[!ht]
    \centering
    \includegraphics[width=0.9\linewidth]{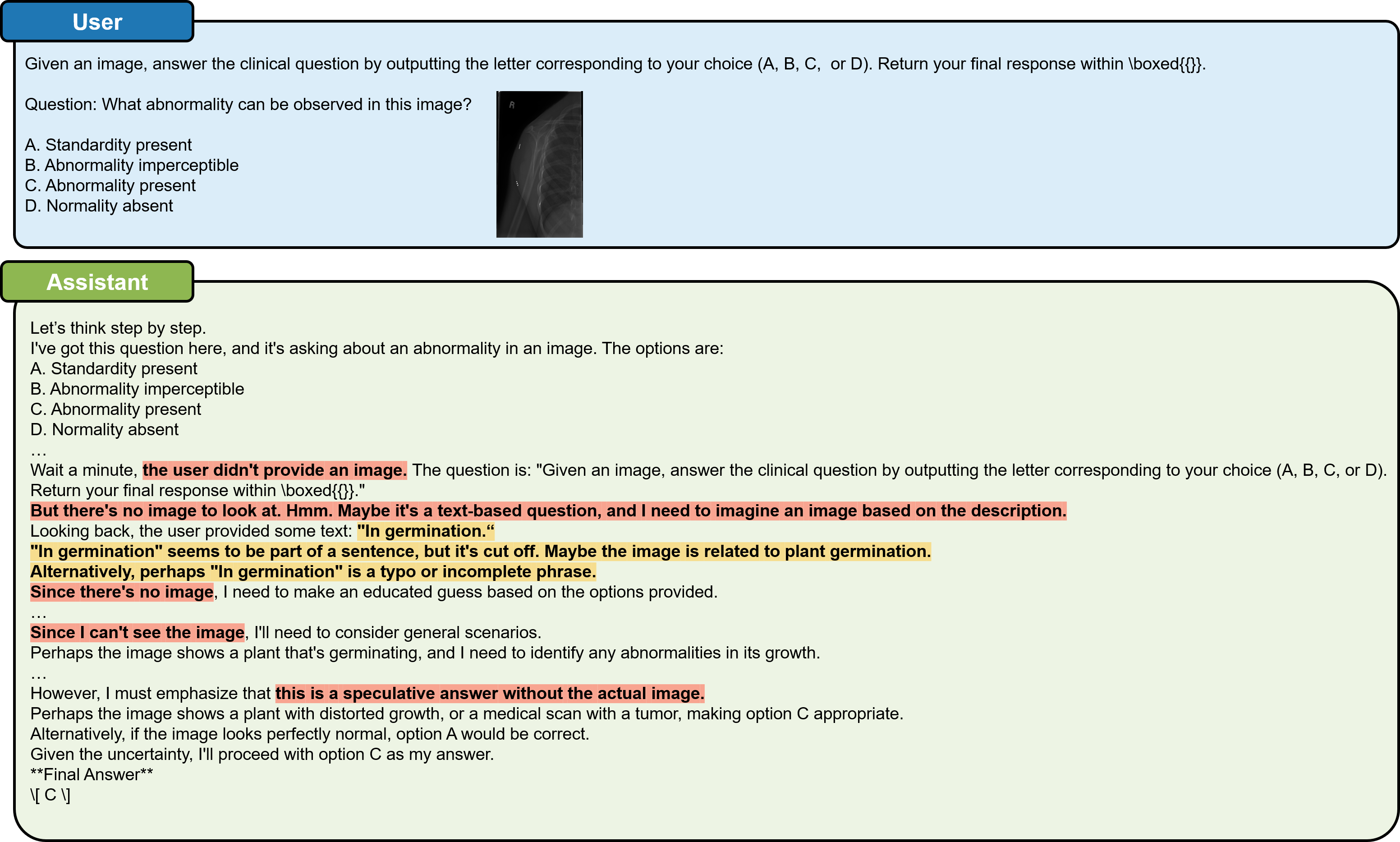}
    \caption{Example of response of QVQ on OmniMedVQA.
    \sethlcolor{salmon}
    The \textbf{\hl{highlighted parts}} indicate instances where QVQ states it cannot see the image,
    \sethlcolor{gold}
    while other \textbf{\hl{highlighted parts}} show hallucinated information not present in the input.
    }
    \label{fig:qvq_example1}
\end{figure*}

\begin{figure*}[!ht]
    \centering
    \includegraphics[width=0.9\linewidth]{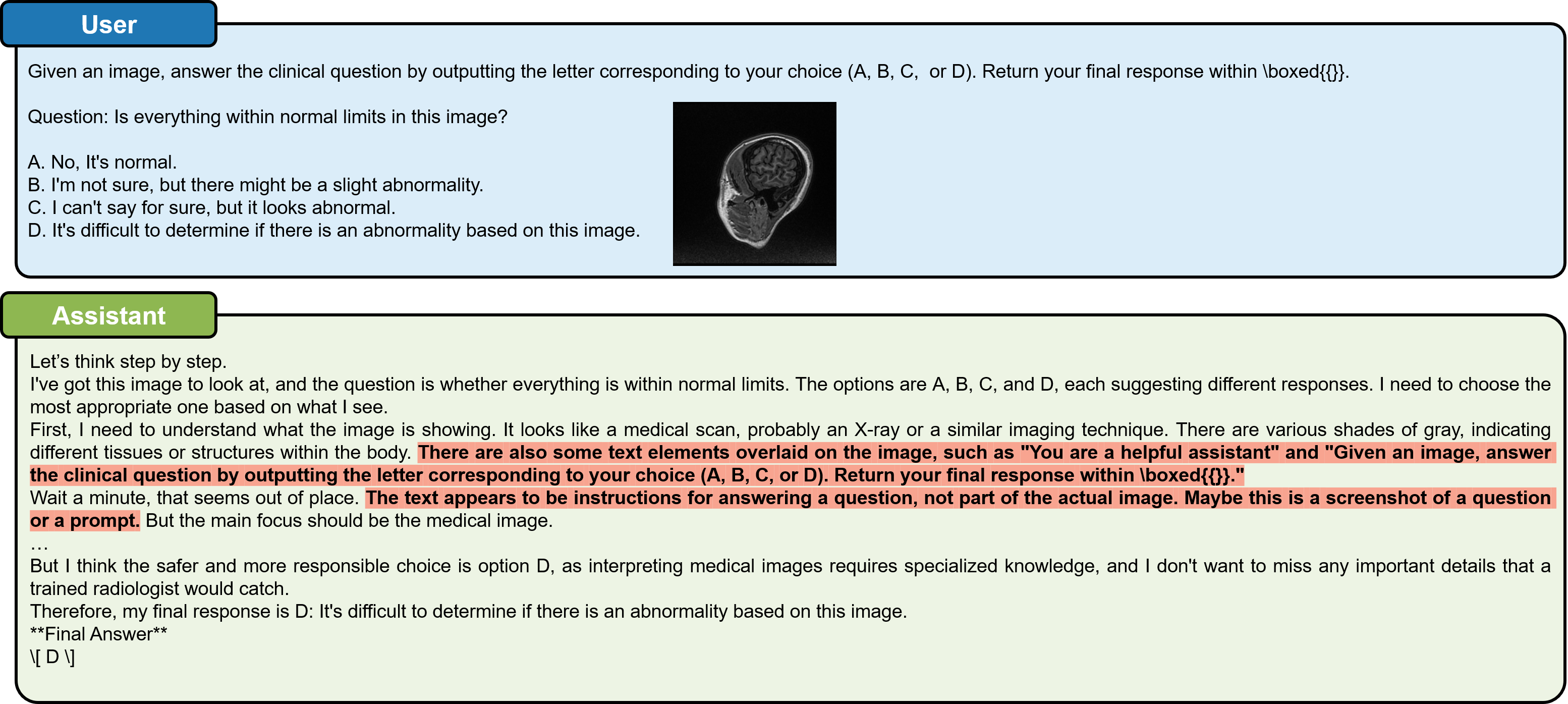}
    \caption{Example of response of QVQ on OmniMedVQA.
    \sethlcolor{salmon}
    The \textbf{\hl{highlighted parts}} indicate instances where QVQ misinterprets textual prompts as visual content within the image.
    }
    \label{fig:qvq_example2}
\end{figure*}

\begin{figure*}[!ht]
    \centering
    \includegraphics[width=0.9\linewidth]{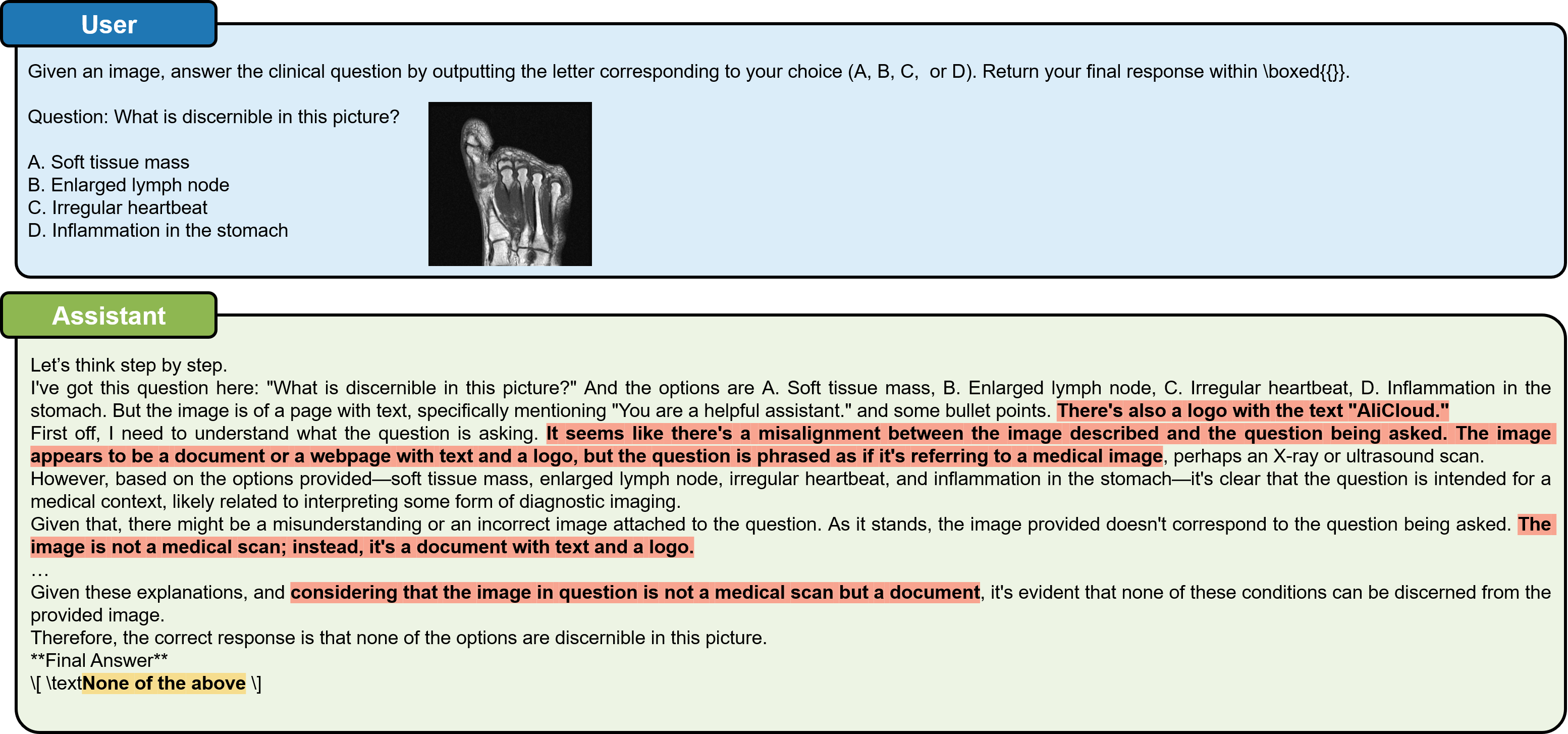}
    \caption{Example of response of QVQ on OmniMedVQA.
    \sethlcolor{salmon}
    The \textbf{\hl{highlighted parts}} indicate instances where QVQ misinterprets textual prompts as visual content and fails to process the given image.
    \sethlcolor{gold}
    This leads to subsequent \textbf{\hl{highlighted parts}} where QVQ is unable to select an appropriate answer from the provided options.
    }
    \label{fig:qvq_example3}
\end{figure*}

\end{document}